\documentclass[letterpaper]{article}
\usepackage{aaai2026}
\usepackage{times}
\usepackage{helvet}
\usepackage{courier}
\usepackage[hyphens]{url}
\usepackage{graphicx}
\usepackage{natbib}
\usepackage{caption}
\usepackage{algorithm}
\usepackage{algpseudocode}
\usepackage{newfloat}
\DeclareCaptionStyle{ruled}{labelfont=normalfont,labelsep=colon,strut=off}
\floatstyle{ruled}
\usepackage{amsmath}
\usepackage{tabularx}
\usepackage{array}
\usepackage{booktabs}
\usepackage{amsfonts}
\usepackage{multirow}
\usepackage{makecell} 
\title{D$^2$HScore: Reasoning-Aware Hallucination Detection via Semantic Breadth and Depth Analysis in LLMs}

\author{
Yue Ding\textsuperscript{1,2,\ddag,*}, 
Xiaofang Zhu\textsuperscript{3,4,\ddag}, 
Tianze Xia\textsuperscript{2,\ddag}, 
Junfei Wu\textsuperscript{1,2}, 
Xinlong Chen\textsuperscript{1,2}, \\
Qiang Liu\textsuperscript{1,2,\dag}, 
Liang Wang\textsuperscript{1,2} \\
}

\affiliations{
\textsuperscript{1}NLPR, Institute of Automation, Chinese Academy of Sciences (CASIA) \\
\textsuperscript{2}School of Artificial Intelligence, University of Chinese Academy of Sciences (UCAS) \\
\textsuperscript{3}	Computer Science, Tsinghua University \\
\textsuperscript{4}	RealAI \\
\dag Corresponding author: qiang.liu@nlpr.ia.ac.cn \\
\ddag Equal contribution.
}
\makeatletter
\copyrighttext{*Work done during an internship at RealAI.} 
\makeatother

\begin{document}

\maketitle

\begin{abstract}
Although large Language Models (LLMs) have achieved remarkable success, their practical application is often hindered by the generation of non-factual content, which is called "hallucination". Ensuring the reliability of LLMs' outputs is a critical challenge, particularly in high-stakes domains such as finance, security, and healthcare. In this work, we revisit hallucination detection from the perspective of model architecture and generation dynamics. Leveraging the multi-layer structure and autoregressive decoding process of LLMs, we decompose hallucination signals into two complementary dimensions: the semantic breadth of token representations within each layer, and the semantic depth of core concepts as they evolve across layers. Based on this insight, we propose \textbf{D$^2$HScore (Dispersion and Drift-based Hallucination Score)}, a training-free and label-free framework that jointly measures: (1) \textbf{Intra-Layer Dispersion}, which quantifies the semantic diversity of token representations within each layer; and (2) \textbf{Inter-Layer Drift}, which tracks the progressive transformation of key token representations across layers. To ensure drift reflects the evolution of meaningful semantics rather than noisy or redundant tokens, we guide token selection using attention signals. By capturing both the horizontal and vertical dynamics of representation during inference, D$^2$HScore provides an interpretable and lightweight proxy for hallucination detection. Extensive experiments across five open-source LLMs and five widely used benchmarks demonstrate that D$^2$HScore consistently outperforms existing training-free baselines.
\end{abstract}
\begin{figure*}[t]
    \centering
    \includegraphics[width=\linewidth]{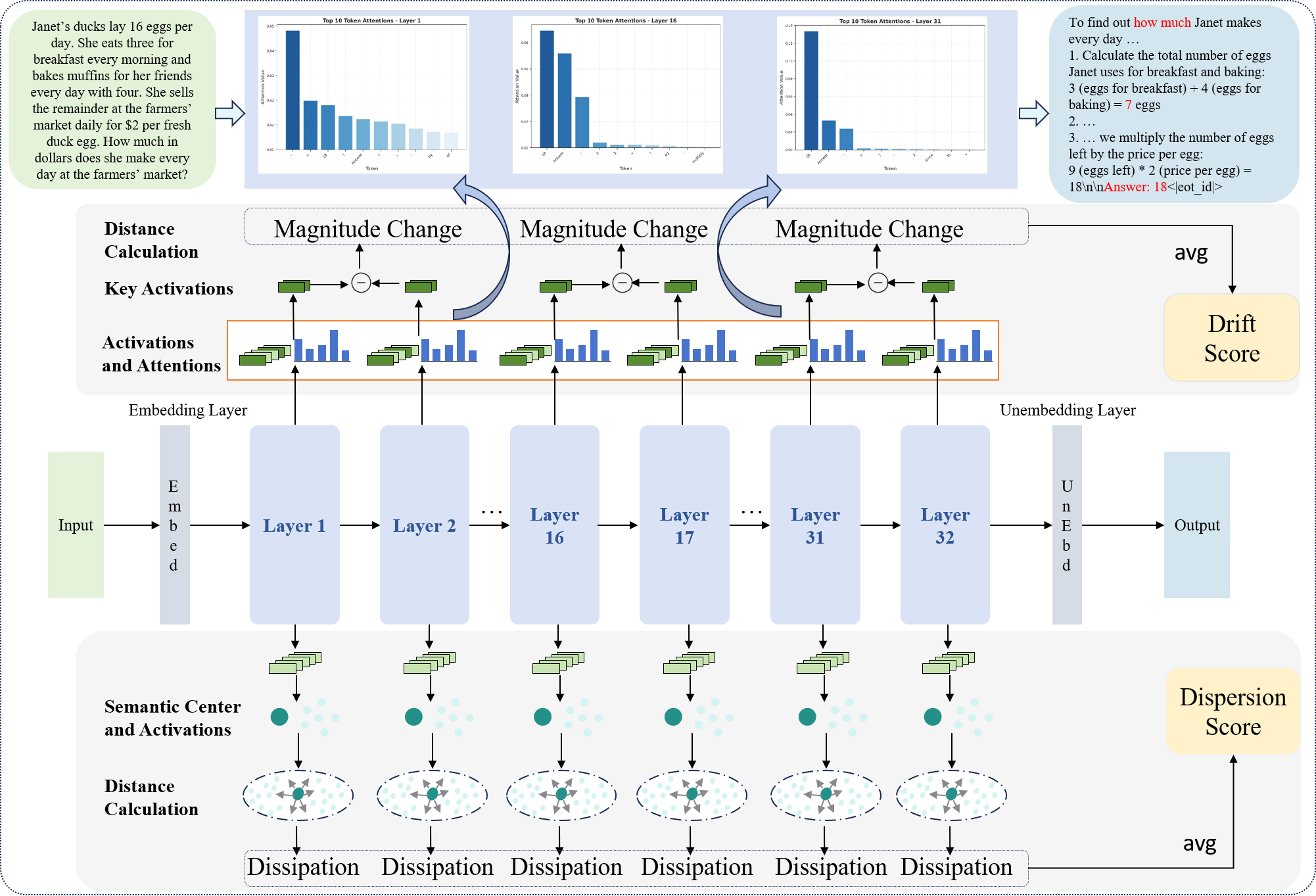}
    \caption{Illustration of \textbf{D$^2$HScore}. Given an input question, we extract the LLM's internal hidden states to compute two complementary scores. The \textbf{Intra-Layer Dispersion Score} (bottom) quantifies the semantic breadth among all token embeddings within a layer, with greater dispersion implying more faithful responses. The \textbf{Inter-Layer Drift Score} (top) measures the semantic depth of key tokens across transformer layers, with larger shifts indicating more coherent reasoning. Hallucinated outputs tend to exhibit lower dispersion and weaker inter-layer dynamics.}
    \label{fig:workflow}
\end{figure*}

\section{Introduction}
Large Language Models (LLMs) have demonstrated impressive capabilities across a wide range of natural language processing tasks, including complex reasoning~\cite{chen2023theoremqa, openai2024openaio1card} and creative text generation~\cite{wu2025writingbench}. These strengths have led to their widespread adoption in real-world applications~\cite{zhao2023survey}. However, a persistent challenge remains: \textit{hallucination}—the generation of fluent yet factually incorrect content~\cite{zhang2023siren}. This issue undermines the reliability of LLMs, particularly in high-stakes domains such as finance, healthcare, and law.

Existing hallucination detection methods primarily focus on either the model's semantic output or its internal representation dynamics, but rarely both. Black-box approaches—such as consistency-based methods~\cite{manakul2023selfcheckgpt, zhang2023sac3, chen2024inside} and tool-assisted verification~\cite{cheng2024small}—evaluate hallucination from the final response alone, often leveraging semantic-level signals like factual inconsistency or variability across generations. While broadly applicable, these methods primarily capture \textit{semantic breadth} and overlook the model’s internal reasoning trace. On the other hand, recent white-box methods attempt to analyze representational patterns inside large language models (LLMs). For instance, Chain-of-Embedding~\cite{wang2025latentspacechainofembeddingenables} proposes to track sentence-level representations across layers by averaging token embeddings, which probes into the model’s \textit{semantic depth} but assumes homogeneity across tokens—a flawed assumption given the diverse roles tokens play in generation. None of these prior methods jointly consider both dimensions.

To bridge this gap, we introduce a new perspective: hallucination arises from a breakdown in either the \textit{semantic breadth} or the \textit{semantic depth} of internal representations—or both. A faithful response should demonstrate a rich and distributed semantic structure within each layer (semantic breadth), and a progressive refinement of key representations across layers (semantic depth). Building on this insight, we propose D$^2$HScore (Dispersion and Drift-based Hallucination Score), the first label-free and training-free framework that jointly captures both dimensions of representational failure.

Specifically, we define:

\begin{itemize}
    \item Semantic Breadth, operationalized via the \textit{Intra-Layer Dispersion Score}, measures how semantically diverse the token representations are within a single layer. Specifically, we assess the spatial spread of embeddings to quantify the richness of semantic unfolding. Lower dispersion often correlates with collapsed, low-information representations characteristic of hallucinated outputs.
    \item Semantic Depth, captured by the \textit{Inter-Layer Drift Score}, measures how the model refines key token representations across layers. We use attention weights to identify semantically important tokens and track their representational trajectories throughout the forward pass. A strong, coherent drift indicates consistent reasoning; weak or stagnant trajectories suggest semantic inconsistency and potential hallucination.
\end{itemize}

By jointly modeling these two aspects of representational dynamics, D$^2$HScore offers a comprehensive and interpretable approach to hallucination detection that moves beyond surface-level heuristics and taps into the reasoning process of LLMs.

Our main contributions are summarized as follows:
\begin{itemize}
    \item We present the first work that analyzes hallucination signals from two complementary perspectives: semantic breadth within layers and semantic depth across layers, offering a new lens on how hallucinations arise in LLMs.
    \item Correspondingly, we design two novel metrics: an intra-layer dispersion score and an attention-guided inter-layer drift score, which together form the core of our unified, training-free framework D$^2$HScore.
    \item We conduct extensive experiments on five open-source LLMs across five diverse benchmarks, showing that D$^2$HScore consistently outperforms existing white-box methods in hallucination detection.
\end{itemize}

\section{Related Work}
The task of detecting hallucinations in Large Language Models (LLMs) is crucial for ensuring their reliability and safety~\cite{zhang2023siren}. Prior studies can be broadly categorized into black-box and white-box approaches. Due to space constraints, we defer a detailed discussion of black-box methods and early white-box techniques based on output probabilities to the Appendix. We focus here on the most relevant line of work: methods that analyze the internal hidden states of LLMs.

\subsection{Hidden states-based Hallucination Detection}
White-box methods leverage the internal representations of LLMs to detect hallucinations, typically offering greater efficiency as they often require only a single forward pass.

Early efforts in this domain utilized supervised classifiers trained on hidden states to detect specific errors like factual inconsistencies, such as ITI~\cite{li2023inference} and MIND~\cite{su2024unsupervised}. While effective for their target tasks, these approaches depend on labeled data and can struggle to generalize to out-of-distribution (OOD) examples or new types of errors.

To overcome these limitations, recent research has shifted towards training-free, unsupervised analysis of hidden representations. Most relevant to our work is the Chain-of-Embedding (CoE) framework~\cite{wang2025latentspacechainofembeddingenables}, which models the geometric trajectory of hidden states across layers. CoE computes the mean representation of output tokens at each layer and tracks changes in this trajectory's magnitude and direction to assess semantic stability.

\section{Method}
\label{sec:method}
Large Language Models (LLMs) generate responses token-by-token, requiring internal representations to maintain sufficient \textbf{semantic breadth} and \textbf{semantic depth}. We hypothesize that hallucinations arise when token representations lack diversity within a layer (breadth) or fail to evolve across layers (depth). To capture both dimensions, we propose \textbf{D$^2$HScore}, which evaluates representational quality via two complementary signals: \textit{Intra-Layer Dispersion} and \textit{Inter-Layer Drift}, as illustrated in Figure~\ref{fig:workflow}.

\subsection{Preliminaries: LLM Forward Propagation}
A Large Language Model (LLM) generates a response auto-regressively. Given an input prompt $P = \{p_1, p_2, \dots, p_m\}$, the model produces an output sequence $Y = \{y_1, y_2, \dots, y_T\}$. During the generation of the $t$-th token $y_t$, the model computes a hidden state representation $h_{l}^{t} \in \mathbb{R}^{d}$ for each layer $l \in \{1, \dots, L\}$, where $L$ is the total number of Transformer layers and $d$ is the hidden dimension.

\begin{equation}
h_{l}^{t} = \text{Transformer}(h_{l-1}^{1}, \dots, h_{l-1}^{t}),\, l=1,\dots,L
\end{equation}

These hidden states are refined progressively through the LLM's stacked Transformer architecture, in which each layer incrementally updates token representations. At the core of each layer lies the multi-head self-attention (MHA) mechanism, which allows the model to compute token-wise contextualized representations by attending to other tokens in the sequence.

The attention computation for each head $h$ is defined as:
\begin{align}
\text{head}_h &= \text{Attention}(Q_h, K_h, V_h) \notag \\
              &= \text{softmax}\left(\frac{Q_h K_h^\top}{\sqrt{d_k}}\right)V_h
\end{align}

The outputs of all heads are then concatenated and linearly projected to form the final MHA output:
\begin{equation}
\text{MHA}(Q, K, V) = \text{Concat}(\text{head}_1, \dots, \text{head}_H)W^O
\end{equation}

By leveraging the attention scores computed within the model, we can identify which tokens contribute most prominently to the model’s reasoning. These high-attention tokens often carry the core semantics of the generated response and serve as the foundation for our subsequent drift analysis.

\subsection{Semantic Breadth: Intra-Layer Dispersion}
The \textit{Dispersion Score} quantifies the semantic diversity of token representations within a single layer. Since LLMs generate tokens autoregressively, a faithful response should exhibit semantically rich and varied representations as tokens contribute unique information. In contrast, hallucinated responses tend to show overly uniform representations, reflecting semantic collapse.

\begin{figure}[h!]
    \centering
    \includegraphics[width=1.0\linewidth]{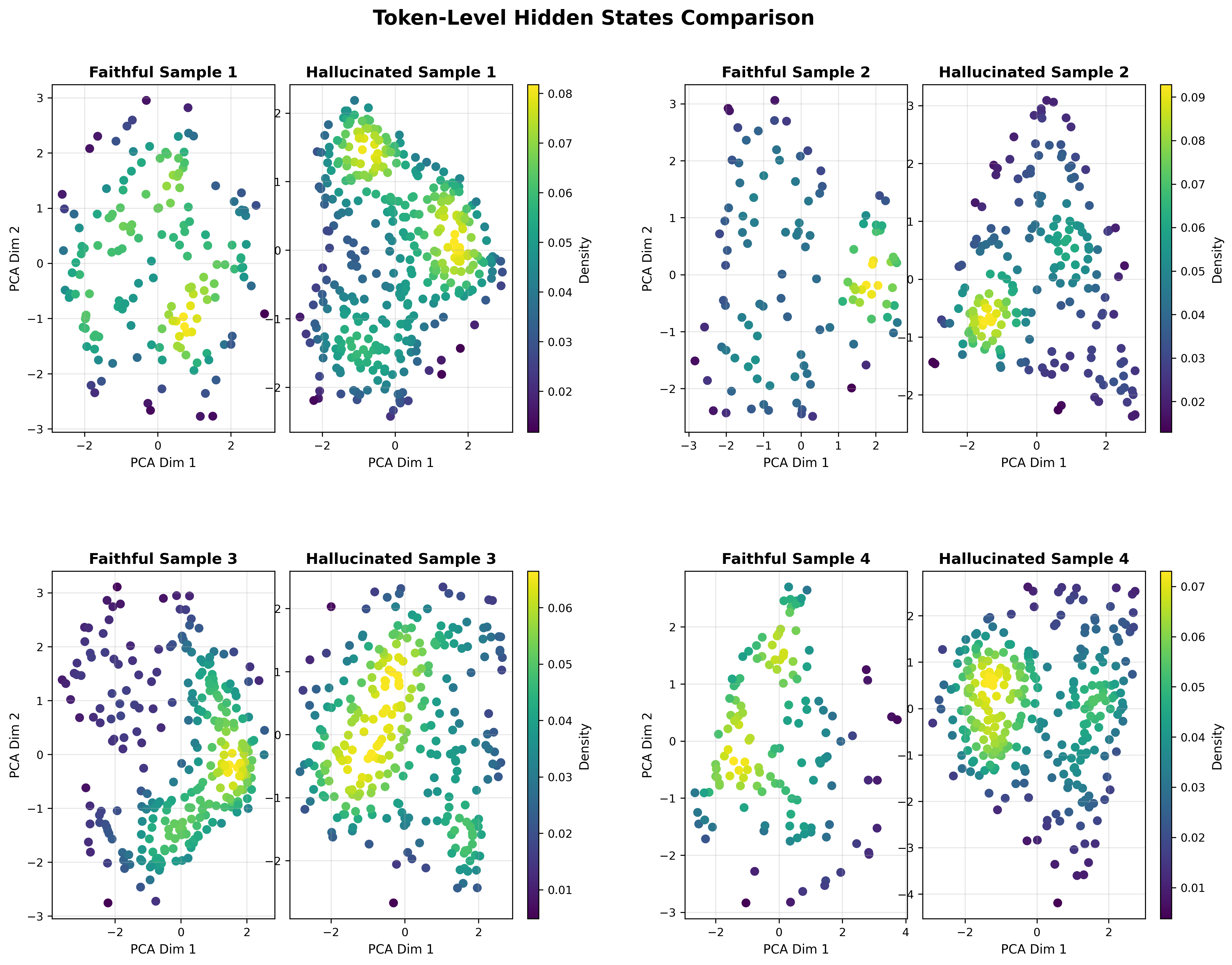} 
    \caption{Visualization of token-level hidden states at layer 10 from the Llama3.1-8B model on GSM8K. Each subplot compares a faithful (left) and hallucinated (right) sample using PCA. Colors represent token density.}
    \label{fig:pca_dispersion}
\end{figure}

We visualize token-level hidden states from layer 10 of the Llama3.1-8B model on the GSM8K dataset. Figure~\ref{fig:pca_dispersion} shows that faithful responses (left) exhibit widely spread token embeddings, while hallucinated ones (right) form tight clusters. This suggests that low dispersion is associated with hallucinated content. The specific examples used in this analysis are detailed in the Appendix.

Based on the above analysis, we now formally define the Dispersion Score.

First, for each layer $l$, we determine the semantic center $c_l$ of all $T$ tokens in the response:
\begin{equation}
c_l = \frac{1}{T} \sum_{t=1}^{T} h_l^t
\end{equation}
Next, the dispersion for layer $l$, denoted $D_l$, is computed based on the mean L2 distance:
\begin{equation}
D_l = \frac{1}{T} \sum_{t=1}^{T} ||h_l^t - c_l||_2
\end{equation}
Finally, the overall Dispersion Score for the entire response is obtained by averaging the dispersion values across all $L$ layers of the model:
\begin{equation}
\text{Score}_{\text{Dispersion}} = \frac{1}{L} \sum_{l=1}^{L} D_l    
\end{equation}

\begin{figure}[h!]
    \centering
    \includegraphics[width=1.0\linewidth]{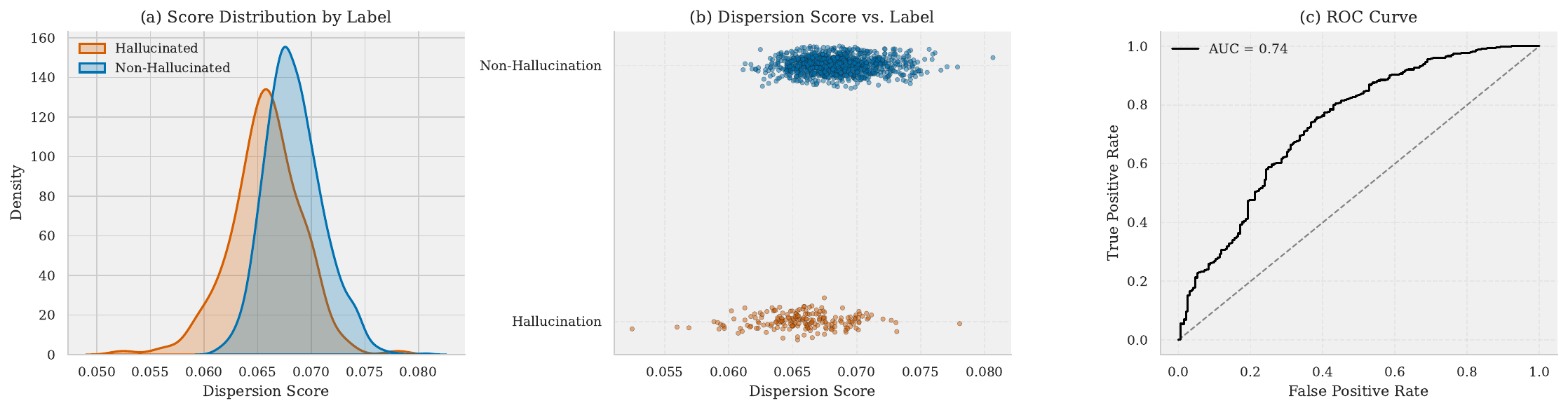} 
    \caption{Distribution of Dispersion Scores on the GSM8K dataset of Llama-3.1-8B model.  (a) Distribution of dispersion scores. (b) Scatter plot of dispersion scores. (c) ROC curve of the dispersion score, achieving an AUC of 0.74.}
    \label{fig:dispersion_dist}
\end{figure}

The results in figure~\ref{fig:dispersion_dist} validates our Dispersion Score on the GSM8K benchmark, demonstrating that hallucinated responses consistently yield lower scores. This clear separation is visible in the distribution and scatter plots (a, b), and is quantified by the ROC analysis (c), which achieves an AUC of 0.74. This confirms low semantic dispersion is a reliable indicator of hallucination.

\subsection{Semantic Depth: Inter-Layer Drift}
\label{sub:drift}
The \textit{Drift Score} measures the degree to which key token representations evolve across layers. This reflects the model’s ability to progressively refine core semantics, a key characteristic of faithful reasoning supported by its stacked Transformer architecture.

To identify semantically important tokens, we first identify a set of key tokens $K_l$ at each layer $l$ by focusing on the attention distribution from the final generated token ($T$). Specifically, we average the attention matrices $A_l^{(h)}$ across all $H$ heads and define the importance score $s_{l,j}$ for each token $j$ as the attention it receives from the final token:
\begin{equation}
\label{eq:importance_score}
s_{l,j} = \left( \frac{1}{H} \sum_{h=1}^{H} A_l^{(h)} \right)_{Tj}
\end{equation}
The key token set $K_l$ is then formed by selecting the top $k$ percent of tokens based on these scores.

Next, we compute the layer's core representation $\bar{h}_l$ using only these key tokens:
\begin{equation}
\label{eq:core_rep}
\bar{h}_l = \frac{1}{|K_l|} \sum_{t \in K_l} h_l^t
\end{equation}

Finally, the Drift Score is defined as the average L2 distance between the core representations of adjacent layers. This captures the magnitude of ``semantic drift'' as the model processes the information more deeply:
\begin{equation}
\label{eq:drift_score}
\text{Score}_{\text{Drift}} = \frac{1}{L-1} \sum_{l=1}^{L-1} ||\bar{h}_{l+1} - \bar{h}_l||_2
\end{equation}

\begin{figure}[h!]
    \centering
    \includegraphics[width=1.0\linewidth]{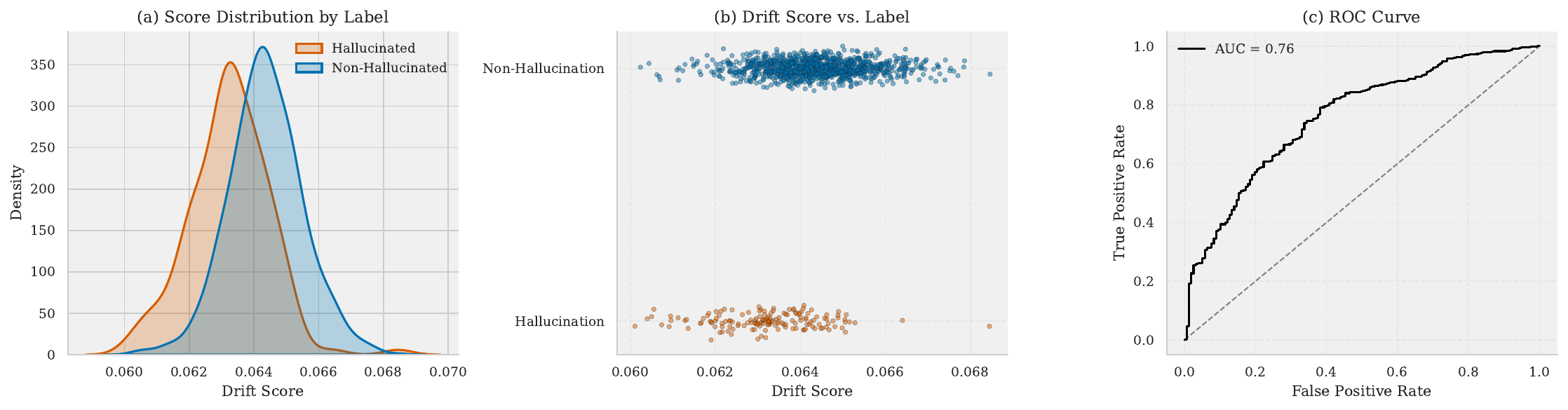} 
    \caption{Distribution of Drift Scores on the GSM8K dataset of Llama-3.1-8B model. (a) Distribution of drift scores. (b) Scatter plot of drift scores. (c) ROC curve of the drift score, achieving an AUC of 0.76.}
    \label{fig:drift_dist}
\end{figure}

Figure~\ref{fig:drift_dist} illustrates that hallucinated responses yield consistently lower Drift Scores on GSM8K. The distributional differences are shown in plots (a, b), and ROC analysis in (c) achieves an AUC of 0.76, confirming that limited representational evolution is indicative of hallucination.

\subsection{The D$^2$HScore: Fusion of Breadth and Depth}
Finally, we combine the two complementary metrics into a single, robust score, the D²HScore. To be noted, both the Dispersion Score and the Drift Score are first normalized to ensure a fair contribution to our final D$^2$HScore. We then perform a linear summation:
\begin{multline}
\text{D$^2$HScore} = w_1 \cdot \text{Normalize}(\text{Score}_{\text{Dispersion}}) \\
+ w_2 \cdot \text{Normalize}(\text{Score}_{\text{Drift}})
\end{multline}

In our implementation, we assign equal weights to the two components, setting $w_1 = w_2 = 0.5$. This balanced fusion enables D$^2$HScore to jointly exploit intra-layer semantic diversity and inter-layer representational dynamics, offering a more comprehensive and robust estimation of response factuality.
\begin{table*}[ht]
\centering
\begin{tabular}{cl|ccc|ccc}
\toprule
\multirow{2}{*}{Model} & \multirow{2}{*}{Method} & 
\multicolumn{3}{c|}{GSM8K} & 
\multicolumn{3}{c}{TheoremQA} \\
\cmidrule(lr){3-5} \cmidrule(lr){6-8}
& & AUROC$\uparrow$ & FPR$\downarrow$ & AUPR$\uparrow$ & 
AUROC$\uparrow$ & FPR$\downarrow$ & AUPR$\uparrow$ \\
\midrule

\multirow{9}{*}{Llama2-7B} 
& Maxprob & 57.96 & 93.37 & 30.71 & 51.37 & 89.61 & 12.58 \\
& ppl & 57.88 & 93.67 & 30.47 & 51.77 & 89.91 & 12.77 \\
& entropy & 56.68 & 92.65 & 31.75 & 51.05 & 91.99 & 12.63 \\
& CoE-R & 61.41 & 90.82 & 33.48 & 58.94 & 74.04 & 12.58 \\
& CoE-C & 53.76 & 92.86 & 29.30 & 61.42 & 75.22 & 15.24 \\
& Temp.Scaling & 57.29 & 93.47 & 30.01 & 53.69 & 93.29 & 14.76 \\
& Energy & 57.73 & 92.76 & 31.15 & 52.53 & 89.43 & 12.77 \\
& D2HScore & \textbf{62.16} & \textbf{90.51} & \textbf{33.81} & \textbf{63.17} & \underline{74.86} & \textbf{19.73} \\
\midrule

\multirow{9}{*}{Qwen1.5-7B} 
& Maxprob & 68.98 & 77.95 & 76.94 & 68.13 & 95.17 & 40.55 \\
& ppl & 69.33 & 77.76 & 77.15 & 68.77 & 94.44 & 40.93 \\
& entropy & 70.90 & 74.80 & 78.22 & 68.38 & 95.75 & 40.05 \\
& CoE-R & 43.14 & 95.28 & 56.14 & 62.97 & 95.90 & 34.46 \\
& CoE-C & 71.08 & 84.25 & 78.73 & 71.39 & 85.65 & 39.06 \\
& Temp.Scaling & 67.91 & 79.33 & 76.18 & 68.19 & 93.85 & 40.69 \\
& Energy & 73.69 & 73.03 & 80.82 & 67.30 & 90.63 & 29.95 \\
& D2HScore & \textbf{74.48} & \textbf{81.10} & \textbf{82.46} & \textbf{75.88} & \underline{89.31} & \textbf{48.42} \\
\midrule

\multirow{9}{*}{DS-Llama-8B}
& Maxprob & 64.12 & 83.16 & 85.93 & 54.40 & 93.73 & 22.36 \\
& ppl & 64.25 & 81.75 & 85.96 & 55.31 & 93.57 & 22.96 \\
& entropy & 65.20 & 82.11 & 86.27 & 54.93 & 94.04 & 22.98 \\
& CoE-R & 57.16 & 90.53 & 80.78 & 47.28 & 96.24 & 19.22 \\
& CoE-C & 64.53 & 80.00 & 85.01 & 59.56 & 87.93 & 30.24 \\
& Temp.Scaling & 63.55 & 83.51 & 85.72 & 54.85 & 94.20 & 22.52 \\
& Energy & 64.35 & 86.32 & 85.15 & 49.41 & 95.77 & 19.47 \\
& D2HScore & \textbf{65.30} & \textbf{80.00} & 85.35 & \textbf{66.83} & \textbf{84.01} & \textbf{35.68} \\
\midrule

\multirow{9}{*}{DS-Qwen-7B}
& Maxprob & 63.30 & 84.62 & 90.07 & 49.70 & 93.32 & 23.99 \\
& ppl & 63.66 & 84.62 & 90.27 & 50.57 & 93.82 & 24.50 \\
& entropy & 64.58 & 82.42 & 90.51 & 50.19 & 92.15 & 24.33 \\
& CoE-R & 67.67 & 80.22 & 91.59 & 51.77 & 94.66 & 25.28 \\
& CoE-C & 66.99 & 81.32 & 91.36 & 51.08 & 94.82 & 24.94 \\
& Temp.Scaling & 62.76 & 86.81 & 89.96 & 50.18 & 93.49 & 24.23 \\
& Energy & 53.90 & 91.21 & 86.03 & 41.76 & 96.66 & 21.08 \\
& D2HScore & \textbf{74.18} & \textbf{78.57} & \textbf{94.08} & \textbf{76.43} & \textbf{82.80} & \textbf{56.98} \\
\bottomrule
\end{tabular}
\caption{AUROC, FPR95, and AUPR results on GSM8K and TheoremQA with different LLMs.}
\label{tab:gsm8k_theoremqa_results}
\end{table*}
\begin{table*}[ht]
\centering
\begin{tabular}{cl|ccc|ccc}
\toprule
\multirow{2}{*}{Model} & \multirow{2}{*}{Method} & 
\multicolumn{3}{c|}{MMLU} & 
\multicolumn{3}{c}{Belebele} \\
\cmidrule(lr){3-5} \cmidrule(lr){6-8}
& & AUROC$\uparrow$ & FPR$\downarrow$ & AUPR$\uparrow$ &
AUROC$\uparrow$ & FPR$\downarrow$ & AUPR$\uparrow$ \\
\midrule

\multirow{9}{*}{Llama2-7B} 
& Maxprob & 46.70 & 96.44 & 36.60 & 43.93 & 94.78 & 38.01 \\
& ppl & 46.78 & 96.72 & 36.74 & 44.37 & 95.74 & 38.25 \\
& entropy & 47.37 & 96.44 & 37.00 & 43.54 & 95.36 & 37.69 \\
& CoE-R & 62.44 & 83.90 & 48.41 & 60.06 & 88.97 & 54.55 \\
& CoE-C & 56.55 & 84.62 & 42.67 & 42.59 & 95.74 & 37.52 \\
& Temp.Scaling & 46.51 & 96.58 & 36.61 & 44.56 & 94.39 & 38.44 \\
& Energy & 52.26 & 93.30 & 40.92 & 39.45 & 96.71 & 36.63 \\
& D2HScore & \underline{60.17} & 84.90 & \underline{46.95} & \textbf{72.16} & \textbf{71.57} & \textbf{62.77} \\
\midrule

\multirow{9}{*}{Llama3.1-8B}
& Maxprob & 52.85 & 88.12 & 71.97 & 57.02 & 81.69 & 92.55 \\
& ppl & 53.93 & 87.50 & 72.95 & 59.33 & 80.28 & 93.38 \\
& entropy & 56.64 & 86.88 & 75.57 & 64.42 & 81.69 & 94.91 \\
& CoE-R & 67.56 & 76.56 & 80.71 & 77.37 & 66.20 & 97.01 \\
& CoE-C & 67.71 & 71.88 & 81.24 & 69.79 & 64.79 & 94.39 \\
& Temp.Scaling & 59.90 & 85.82 & 80.94 & 55.59 & 84.51 & 91.73 \\
& Energy & 69.97 & 80.97 & 86.56 & 77.86 & 59.15 & 96.77 \\
& D2HScore & \underline{69.70} & \underline{70.94} & \underline{81.85} & \textbf{80.96} & \textbf{53.52} & \textbf{97.39} \\
\midrule

\multirow{9}{*}{Qwen1.5-7B} 
& Maxprob & 53.50 & 92.50 & 57.05 & 44.96 & 93.92 & 77.36 \\
& ppl & 53.64 & 93.08 & 57.48 & 46.07 & 95.03 & 78.07 \\
& entropy & 53.72 & 92.88 & 57.15 & 46.09 & 92.27 & 77.22 \\
& CoE-R & 48.67 & 95.77 & 53.28 & 54.69 & 95.58 & 84.43 \\
& CoE-C & 64.57 & 85.00 & 65.56 & 55.10 & 91.71 & 84.00 \\
& Temp.Scaling & 53.38 & 93.85 & 57.38 & 45.43 & 96.69 & 78.26 \\
& Energy & 56.37 & 90.77 & 60.22 & 46.81 & 92.82 & 77.60 \\
& D2HScore & \textbf{65.17} & \textbf{83.27} & \textbf{66.26} & \textbf{61.27} & \textbf{90.96} & 82.14 \\
\midrule

\multirow{9}{*}{DS-Llama-8B} 
& Maxprob & 88.36 & 27.98 & 72.69 & 92.49 & 20.90 & 96.71 \\
& ppl & 88.39 & 28.55 & 72.53 & 92.37 & 22.60 & 96.68 \\
& entropy & 88.83 & 28.12 & 74.67 & 93.31 & 19.77 & 97.46 \\
& CoE-R & 59.94 & 87.80 & 45.64 & 61.86 & 89.83 & 86.29 \\
& CoE-C & 92.66 & 23.39 & 85.11 & 94.72 & 15.82 & 97.92 \\
& Temp.Scaling & 88.00 & 28.41 & 71.32 & 91.54 & 27.12 & 96.11 \\
& Energy & 91.67 & 21.81 & 81.06 & 94.34 & 16.95 & 97.80 \\
& D2HScore & \textbf{93.08} & \underline{22.24} & \textbf{86.90} & \textbf{98.49} & \textbf{6.74} & \textbf{99.53} \\
\midrule
\bottomrule
\end{tabular}
\caption{AUROC, FPR95, and AUPR results on MMLU and Belebele with different LLMs.}
\label{tab:mmlu_belebele_results}
\end{table*}

\section{Experiments}
\subsection{Setup}
\paragraph{Dataset.}
To comprehensively evaluate our method, we select five diverse benchmarks covering a wide spectrum of tasks. For reasoning, we use GSM8K~\cite{cobbe2021training} for mathematics and TheoremQA~\cite{chen2023theoremqa} for scientific theorems. For broad knowledge and multi-task understanding, we include MMLU~\cite{hendrycks2020measuring}. For reading comprehension, we use the English subset of Belebele~\cite{bandarkar2023belebele}, and for multilingual generalization, we include MGSM~\cite{shi2022language}. Detailed descriptions of each dataset are available in the Appendix.

\paragraph{Model.}
We evaluate our method on five models from two categories: Instruction-Tuned Models (Llama2-7B-Instruct~\cite{touvron2023llama}, Llama3.1-8B-Instruct~\cite{llama3}, and Qwen1.5-7B-Instruct~\cite{qwen15}) and Long Reasoning Models (LRMs) (DeepSeek-distilled-Llama-8B and DeepSeek-distilled-Qwen-7B~\cite{deepseekai2025deepseekr1incentivizingreasoningcapability}).

\paragraph{Baseline.}
We compare our method against seven baselines grouped into three categories. The first includes common uncertainty metrics: Maximum Softmax Probability (MaxProb)~\cite{hendrycks2017baseline}, Perplexity (PPL)~\cite{si2023promptinggpt3reliable}, and Entropy~\cite{Huang_2025}. The second covers calibration-based methods: Temperature Scaling~\cite{shih2023long} and Energy~\cite{liu2020energy}. The third consists of two recent representation-based scores: CoE-R and CoE-C~\cite{wang2025latentspacechainofembeddingenables}. Mathematical formulations for all baselines are provided in Appendix.

\paragraph{Evaluation Metric.}

We frame hallucination detection as a binary classification task, labeling each response as correct or hallucinated via exact match. We evaluate performance using three standard metrics: AUROC~\cite{boyd2013area}, AUPR~\cite{boyd2013area}, and FPR95~\cite{hendrycks2022scaling}. Detailed descriptions of these metrics are provided in Appendix.

\begin{table*}[ht]
\centering
\begin{tabular}{cl|ccc|ccc}
\toprule
\textbf{Model} & \textbf{Method}
& \multicolumn{3}{c|}{\textbf{fr}}
& \multicolumn{3}{c}{\textbf{ja}} \\
\cmidrule(lr){3-5} \cmidrule(lr){6-8}
& & AUROC$\uparrow$ & FPR@95$\downarrow$ & AUPR$\uparrow$
  & AUROC$\uparrow$ & FPR@95$\downarrow$ & AUPR$\uparrow$ \\
\midrule

\multirow{8}{*}{Llama-7B}
& Maxprob        & 47.21 & 95.63 & 17.19 & 57.04 & 90.61 & 16.08 \\
& PPL            & 48.08 & 95.63 & 18.84 & 56.91 & 89.67 & 16.01 \\
& Entropy        & 49.65 & 95.15 & 19.70 & 56.22 & 92.02 & 15.59 \\
& Temp           & 46.71 & 95.15 & 18.59 & 57.42 & 89.67 & 16.49 \\
& Energy         & 47.16 & 92.23 & 15.86 & 49.75 & 96.24 & 14.51 \\
& CoE-R          & 65.75 & 78.64 & 24.22 & 49.17 & 90.61 & 13.86 \\
& CoE-C          & 55.25 & 90.78 & 25.89 & 48.93 & 94.37 & 14.13 \\
& \textbf{D$^2$HScore} & \textbf{71.40} & \textbf{79.13} & \textbf{32.27} & \textbf{60.75} & \textbf{85.92} & \textbf{17.28} \\
\midrule

\multirow{8}{*}{DS-8B}
& Maxprob        & 57.34 & 93.94 & 53.71 & 57.89 & 83.58 & 50.17 \\
& PPL            & 57.58 & 91.67 & 54.16 & 58.97 & 85.07 & 51.39 \\
& Entropy        & 57.27 & 90.91 & 52.64 & 59.71 & 87.31 & 53.68 \\
& Temp           & 56.81 & 91.67 & 53.40 & 56.97 & 84.33 & 49.39 \\
& Energy         & 54.41 & 90.91 & 50.12 & 59.06 & 91.79 & 56.83 \\
& CoE-R          & 56.55 & 94.70 & 52.22 & 66.65 & 85.82 & 60.03 \\
& CoE-C          & 48.93 & 96.21 & 47.14 & 66.63 & 86.57 & 60.76 \\
& \textbf{D$^2$HScore} & \textbf{61.54} & \textbf{96.21} & \textbf{56.55} & \textbf{71.11} & \textbf{76.87} & \textbf{63.09} \\
\bottomrule
\end{tabular}
\caption{AUROC, FPR@95, and AUPR on MGSM dataset across French (fr) and Japanese (ja) for two LLMs.}
\label{tab:mgsm_results}
\end{table*}

\subsection{Main Results}

As shown in Table~\ref{tab:gsm8k_theoremqa_results} and Table~\ref{tab:mmlu_belebele_results}, our proposed D$^2$HScore consistently outperforms all baseline methods across a wide range of datasets and LLMs. On the GSM8K and TheoremQA datasets (Table~\ref{tab:gsm8k_theoremqa_results}), D$^2$HScore achieves the highest AUROC and AUPR scores in nearly all model configurations, indicating strong discriminative ability in detecting hallucinated responses. For example, on TheoremQA, D$^2$HScore improves the AUPR from 12.77 (Maxprob) to 19.73 on Llama2-7B, and from 40.93 (ppl) to 48.42 on Qwen1.5-7B. This demonstrates that our method is particularly effective for mathematical and symbolic reasoning tasks, where hallucinations tend to be more subtle and structure-dependent.

In Table~\ref{tab:mmlu_belebele_results}, D$^2$HScore also achieves the best or competitive results on MMLU and the English subset of Belebele, two benchmarks focused on knowledge-intensive understanding. On MMLU, D$^2$HScore outperforms energy-based and calibration-based baselines in both AUROC and AUPR, e.g., achieving AUPR of 81.85  on Llama3.1-8B compared to 80.94 (Temp.Scaling) and 80.71 (CoE-R). On the English subset of Belebele, D$^2$HScore achieves strong performance, such as 97.39 AUPR score and 53.52 FPR score on Llama3.1-8B, and up to 99.53 AUPR score on DS-Llama-8B.

These results highlight two key advantages of D$^2$HScore. First, it generalizes well across models of varying scales and architectures, including distilled reasoning-specialized variants such as DS-Qwen-7B and DS-Llama-8B. Second, it offers robust performance across both symbolic reasoning and knowledge-intensive reading tasks.

\subsection{Extended Analysis}

\paragraph{Multi-lingual Results on MGSM}
As shown in Table~\ref{tab:mgsm_results}, D$^2$HScore demonstrates strong multilingual generalization on the MGSM benchmark. Across both French (fr) and Japanese (ja) subsets, it consistently outperforms baselines on Llama2-7B and DS-Llama-8B. For instance, on Llama2-7B (fr), its AUROC of 71.40 significantly surpasses both traditional metrics like Entropy (49.65) and recent methods like CoE-R (65.75). This robust performance extends to DS-Llama-8B (ja), where D$^2$HScore achieves a leading AUPR of 63.09, well ahead of competitors such as Entropy (53.68) and CoE-C (60.76). These results confirm that our method reliably identifies hallucinations in challenging cross-lingual settings.

\begin{figure}[h!]
    \centering
    \includegraphics[width=1.0\linewidth]{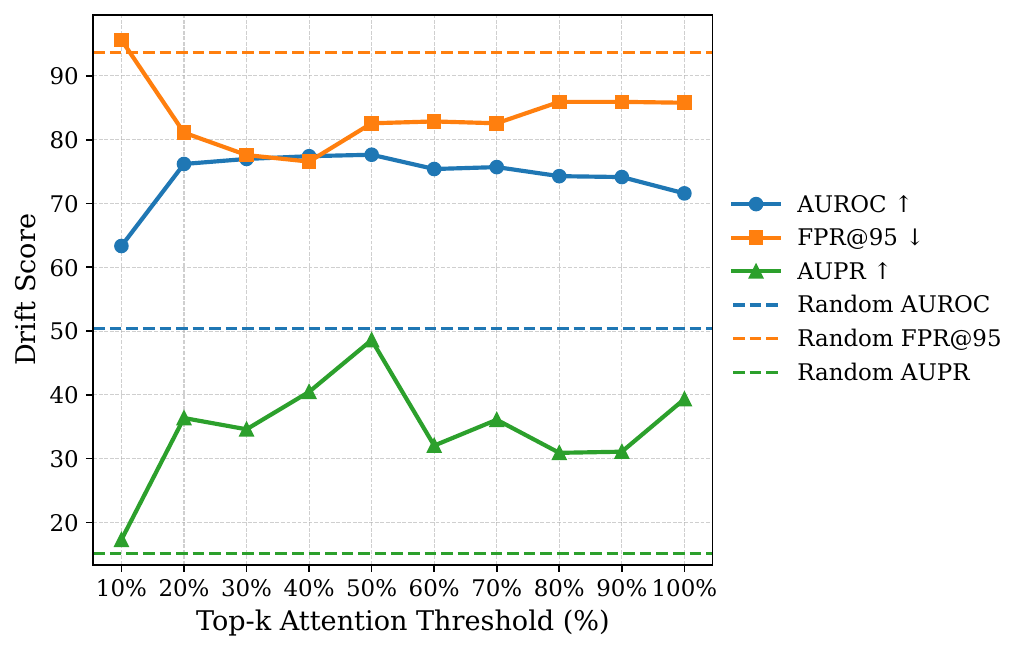} 
    \caption{attention threshold ablation results. The results are from Qwen1.5-7B model on theoremqa dataset.}
    \label{fig:attention_ablation}
\end{figure}

\paragraph{Attention Threshold Ablation}
Figure~\ref{fig:attention_ablation} presents an ablation study on the top-$k$ attention threshold for our Drift Score, conducted on the TheoremQA dataset using the Qwen-1.5-7B model. We test thresholds ranging from 0.1 to 1.0. The results show that our attention-guided key token selection is robust, consistently and significantly outperforming the random results across all metrics (AUROC, FPR@95, and AUPR). While performance is weaker at a very restrictive threshold (k=0.1), it peaks when the threshold is set to 0.4 or 0.5. Notably, this optimal performance is superior to the baseline of using all tokens (k=1.0), which validates our strategy of focusing on a core subset of semantic anchors.

\paragraph{Component Ablation Study}
To assess the contribution of each component, we conduct an ablation study evaluating the Dispersion Score and Drift Score individually. 
As shown in Table~\ref{tab:ablation_results}, the two scores capture complementary aspects of hallucination, with their relative effectiveness varying across models. 
For example, the Drift Score is stronger on Llama-7B, whereas the Dispersion Score is more effective on DS-8B. The combined D$^2$HScore, however, generally yields the best overall performance, demonstrating a clear synergistic effect. 
This is particularly evident on the DS-8B model, where the full score dramatically improves all metrics, reducing the FPR@95 from over 14 to just 6.74.
While minor trade-offs exist in some specific configurations (e.g., FPR@95 on Llama3.1-8B), D$^2$HScore consistently achieves a superior AUROC and AUPR. 
This confirms that both components are crucial for building a robust and comprehensive hallucination detector.

\begin{table}[ht]
\centering
\begin{tabular}{cl|ccc}
\toprule
\textbf{Model} & \textbf{Method} & AUROC↑ & FPR@95↓ & AUPR↑ \\
\midrule
\multirow{3}{*}{Llama-7B} 
& Dispersion & 64.25 & 87.23 & 57.47 \\
& Drift      & 71.55 & 72.34 & 60.98 \\
& D$^2$HScore      & \textbf{72.16} & \textbf{71.57} & \textbf{62.77} \\
\midrule
\multirow{3}{*}{Qwen1.5-7B} 
& Dispersion & 58.31 & 86.19 & 84.20 \\
& Drift      & 59.75 & 87.43 & 83.36 \\
& D$^2$HScore      & \textbf{61.27} & \textbf{90.96} & \textbf{82.14} \\
\midrule
\multirow{3}{*}{Llama3.1-8B} 
& Dispersion & 76.61 & 63.38 & 96.67 \\
& Drift      & 79.25 & \textbf{52.11} & 97.10 \\
& D$^2$HScore      & \textbf{80.96} & 53.52 & \textbf{97.39} \\
\midrule
\multirow{3}{*}{DS-8B} 
& Dispersion & 96.13 & 14.12 & 98.83 \\
& Drift      & 94.83 & 15.25 & 97.97 \\
& D$^2$HScore      & \textbf{98.49} & \textbf{6.74} & \textbf{99.53} \\
\bottomrule
\end{tabular}
\caption{Ablation results comparing Dispersion Score, Drift Score, and D$^2$HScore on four LLMs.}
\label{tab:ablation_results}
\end{table}

\section{Conclusion}
In this paper, we present D$^2$HScore, a novel, training-free framework for hallucination detection in LLMs, grounded in two fundamental dimensions of internal representation: semantic breadth and semantic depth. These are captured via two complementary signals—\textit{Dispersion Score}, which quantifies the intra-layer diversity of token representations, and \textit{Drift Score}, which measures the inter-layer evolution of key semantic tokens. Extensive experiments across models and benchmarks show that this dual-perspective design enables D$^2$HScore to significantly outperform prior white-box methods. Beyond offering an effective and interpretable evaluation tool, our approach opens new directions for hallucination mitigation, such as using D$^2$HScore as a feedback signal to guide generation.

\bibliography{main}

\clearpage
\appendix
\section*{Appendix}

\section*{A. Additional Related Work}

\paragraph{Black-box Hallucination Detection}

Black-box methods assess response quality without access to a model's internal parameters or hidden states. These approaches are especially valuable when working with closed-source models such as GPT-4~\cite{achiam2023gpt}.

One representative paradigm is \textbf{Verbal Confidence (VC)}~\cite{kadavath2022language}, which prompts the LLM to report its own confidence in its generated output. However, VC-based approaches are limited by the tendency of LLMs to be overconfident, often assigning high self-confidence scores even to incorrect or hallucinated outputs.

Another widely adopted strategy is the \textbf{Prompt-Sampling-Aggregation (PSA)} framework~\cite{gao2024spuq}, which perturbs the input prompt (e.g., via rephrasing or stochastic decoding) and evaluates the consistency among multiple generated responses. Methods such as \textbf{SelfCheckGPT}~\cite{manakul2023selfcheckgpt} use sentence-level comparison to identify factual inconsistencies, while $\text{SAC}^3$~\cite{zhang2023sac3} detects hallucinations by analyzing answer consistency across different LLMs or semantically equivalent queries.

Despite their flexibility, black-box approaches present several challenges. VC-based methods struggle with calibration issues and may misrepresent uncertainty. PSA-based methods require multiple inference passes, which significantly increases computational cost and sensitivity to sampling noise. These drawbacks limit their scalability and stability in practical applications.

Given these challenges, recent research has increasingly turned to white-box approaches that leverage internal model representations. Our proposed method, D$^2$HScore, builds on this trend by analyzing both intra-layer semantic dispersion and inter-layer representational drift to detect hallucinations in a training-free and label-free manner.

\paragraph{Output-Probability-Based Methods.}

A prominent class of white-box approaches relies on the final output probability distribution to estimate uncertainty. These include perplexity~\cite{si2023promptinggpt3reliable}, maximum softmax probability, and entropy~\cite{Huang_2025}, which measure how confident the model is in its predicted tokens. Recent extensions like semantic entropy~\cite{kuhn2023semantic} attempt to address limitations by computing uncertainty over semantically distinct completions. However, such methods remain constrained to the final layer and may fail to capture the full trajectory, often being influenced by the model's overconfidence bias.

\section*{B. Pseudocode}
The pseudocode provided in Algorithm~\ref{alg:d2hscore} outlines the computation process of our proposed \textbf{D$^2$HScore}, which assesses the factual consistency of a language model's response based on its internal hidden state dynamics.

\begin{algorithm}[H]
\caption{Compute D\textsuperscript{2}HScore for a Generated Response}
\label{alg:d2hscore}
\begin{algorithmic}[1]
    \Require Hidden states $\{h_l^t \in \mathbb{R}^d\}$ for $l = 1$ to $L$, $t = 1$ to $T$ \\
             Attention matrices $\{A_l \in \mathbb{R}^{T \times T \times H}\}$ per layer \\
             Number of top tokens $k$, weights $w_1$, $w_2$
    \Ensure D\textsuperscript{2}HScore
    
    \Statex \textit{// Dispersion Score Computation}
    \For{$l = 1$ to $L$}
        \State $c_l \gets \frac{1}{T} \sum_{t=1}^{T} h_l^t$ \Comment{Compute semantic center}
        \State $D_l \gets \frac{1}{T} \sum_{t=1}^{T} \| h_l^t - c_l \|_2$ \Comment{Intra-layer dispersion}
    \EndFor
    \State $\text{Score}_{\text{Dispersion}} \gets \frac{1}{L} \sum_{l=1}^{L} D_l$
    
    \Statex
    
    \Statex \textit{// Drift Score Computation}
    \For{$l = 1$ to $L$}
        \State $A_l^{avg} \gets \text{mean}_{h} (A_l^{(h)})$ \Comment{Average over heads}
        \State $s_l \gets \text{mean}_{i}(A_l^{avg}[i, :])$ \Comment{Token importance scores}
        \State $K_l \gets \text{Top-}k\text{ tokens with highest } s_l$
        \State $\bar{h}_l \gets \frac{1}{k} \sum_{t \in K_l} h_l^t$ \Comment{Core representation}
    \EndFor
    \For{$l = 1$ to $L-1$}
        \State $\delta_l \gets \| \bar{h}_{l+1} - \bar{h}_l \|_2$
    \EndFor
    \State $\text{Score}_{\text{Drift}} \gets \frac{1}{L - 1} \sum_{l=1}^{L-1} \delta_l$
    
    \Statex
    
    \Statex \textit{// Final D\textsuperscript{2}HScore}
    \State Normalize $\text{Score}_{\text{Dispersion}}$ and $\text{Score}_{\text{Drift}}$ to $[0, 1]$
    \State $\text{D\textsuperscript{2}HScore} \gets w_1 \cdot \text{NormalizedDispersion} + w_2 \cdot \text{NormalizedDrift}$
    \State \Return D\textsuperscript{2}HScore
\end{algorithmic}
\end{algorithm}

\section*{C. Experiment Details}
\subsection*{Dataset}
\begin{figure}[h!]
    \centering
    \includegraphics[width=1.0\linewidth]{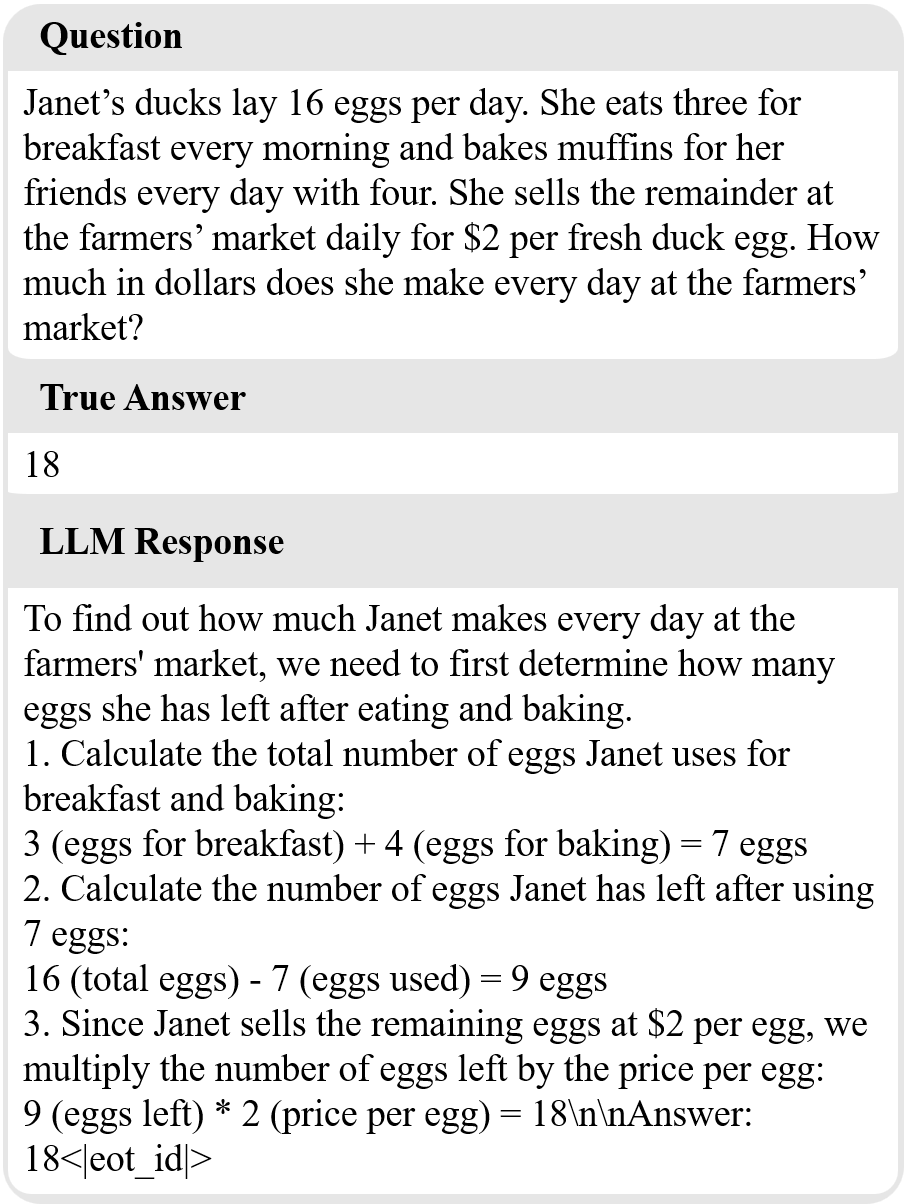} 
    \caption{An example in GSM8K dataset.}
    \label{fig:gsm8k_example}
\end{figure}

\begin{figure}[h!]
    \centering
    \includegraphics[width=1.0\linewidth]{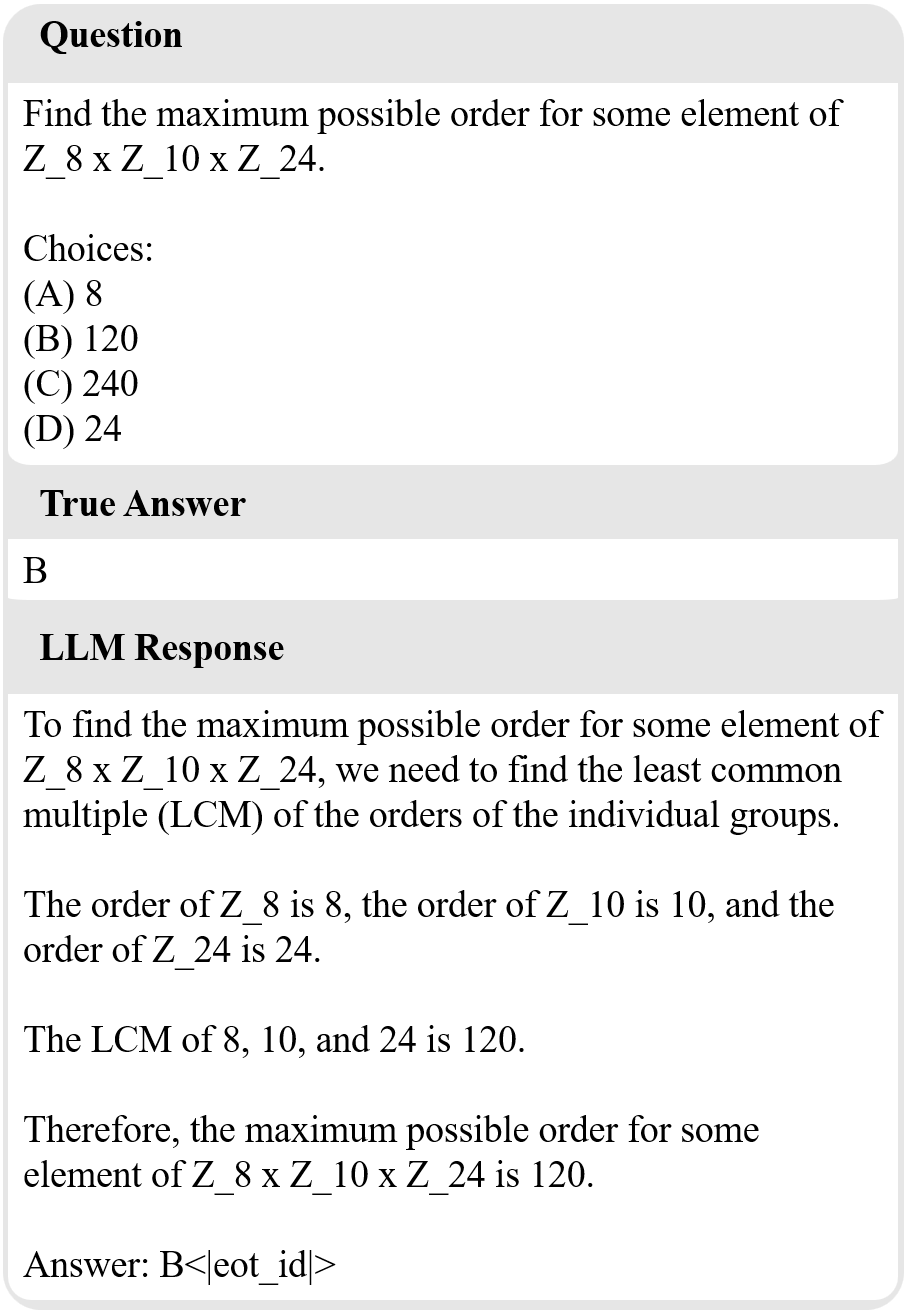} 
    \caption{An example in MMLU dataset.}
    \label{fig:mmlu_example}
\end{figure}

\begin{figure}[h!]
    \centering
    \includegraphics[width=1.0\linewidth]{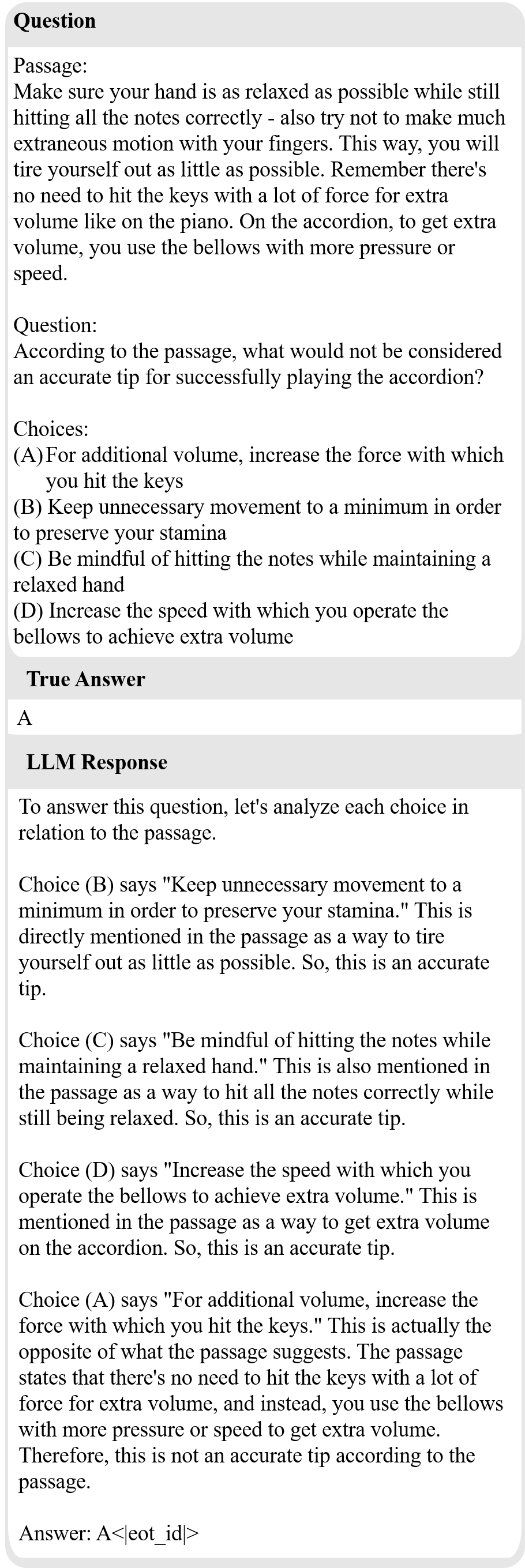} 
    \caption{An example in Belebele dataset.}
    \label{fig:belebele_example}
\end{figure}

\begin{figure}[h!]
    \centering
    \includegraphics[width=1.0\linewidth]{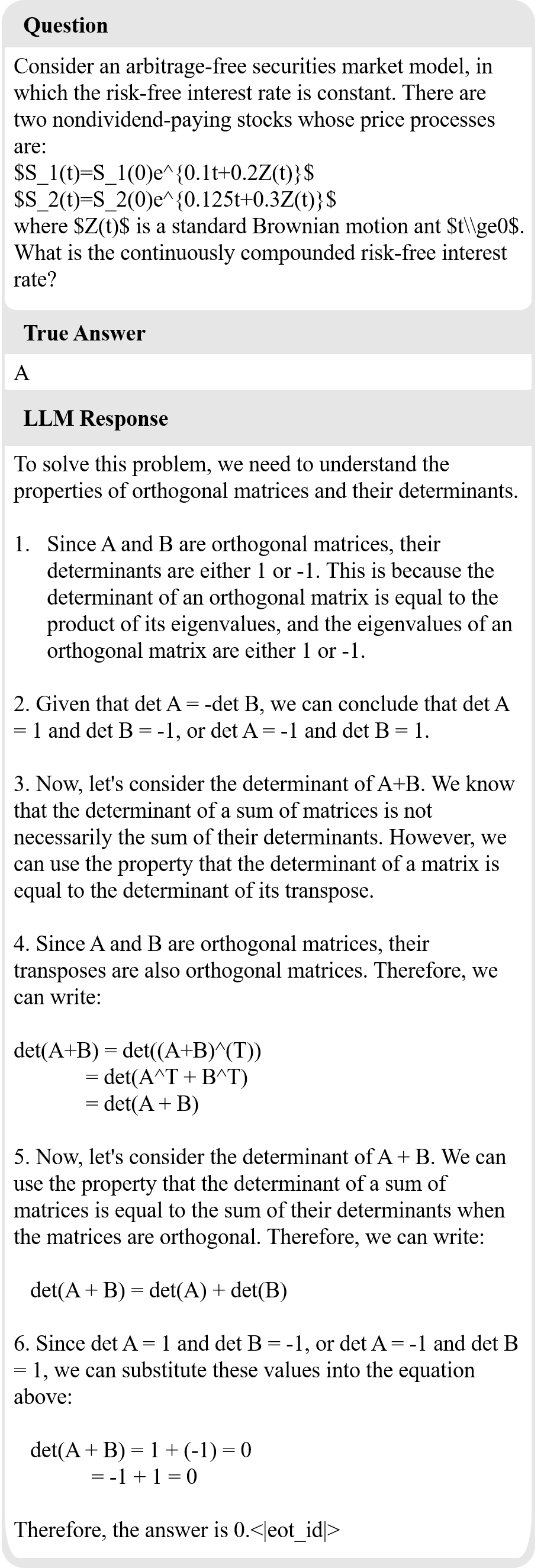} 
    \caption{An example in TheoremQA dataset.}
    \label{fig:theoremqa_example}
\end{figure}

We select five datasets in our main experiment. Figures \ref{fig:gsm8k_example} - \ref{fig:theoremqa_example} are examples in datasets we use.
\begin{itemize}
    \item \textbf{GSM8K}~\cite{cobbe2021training} consists of linguistically diverse math word problems specifically designed for grade school students and created by human writers. It contains 1318 test problems, each requiring 2–8 steps to solve. The challenges primarily involve a sequence of basic arithmetic operations (addition, subtraction, multiplication, and division) to derive the final answer.

    \item \textbf{TheoremQA}~\cite{chen2023theoremqa} consists of various question-answering questions driven by STEM theorems. It contains 800 test problems and covers 350+ theorems spanning across Math, EE\&CS, Physics, and Finance. The dataset is collected by human experts with very high quality.

    \item \textbf{MMLU}~\cite{hendrycks2020measuring} consists of diverse questions that measure knowledge acquired during pretraining. This dataset covers 52 subjects across STEM, the humanities, the social sciences, and more. It ranges in difficulty from an elementary level to an advanced professional level, and can test both world knowledge and problem-solving ability. In our work, we sample 20 examples from each subject, resulting in a 1140-example subset used for evaluation.

    \item \textbf{Belebele}~\cite{bandarkar2023belebele} consists of multilingual, multitask, and multidisciplinary reading comprehension questions, with each language containing 900 article–question pairs. We use the English version of this dataset.

    \item \textbf{MGSM} is a multilingual extension of GSM8K, consisting of math word problems translated into multiple languages. Each problem maintains the original structure and complexity, aiming to evaluate cross-lingual mathematical reasoning abilities.
\end{itemize}

\begin{table}[htbp]
\centering
\begin{tabular}{l l r l l}
\toprule
\textbf{Dataset} & \textbf{Domain} & \textbf{\#Samples} & \textbf{Format} \\
\midrule
GSM8K      & Math         & 1318         & Open QA \\
Belebele   & Understanding& 900          & Open QA \\
MGSM       & Math         & 250/lang     & Open QA \\
TheoremQA  & Reasoning    & 800          & Open QA \\
MMLU       & Knowledge    & 1440         & MCQ \\
\bottomrule
\end{tabular}
\caption{Statistics of datasets used in our evaluation.}
\label{tab:benchmark_simple}
\end{table}

\subsection*{Implementation}
\subsubsection*{Model Source}
All models used in our research are publicly available. 
The specific open-source models, along with their download sources and corresponding licenses, are detailed in Table~\ref{tab:model_details}.

\begin{table*}[ht!]
\centering
\small  
\begin{tabularx}{\textwidth}{>{\raggedright\arraybackslash}p{0.20\textwidth} >{\raggedright\arraybackslash}X >{\raggedright\arraybackslash}X}
\toprule
\textbf{Model Name} & \textbf{Download Link} & \textbf{License Link} \\
\midrule
Llama2-7B-Instruct &
\url{https://huggingface.co/meta-llama/Llama-2-7b-chat-hf} &
\url{https://ai.meta.com/llama/license} \\
\addlinespace
Llama3.1-8B-Instruct &
\url{https://huggingface.co/meta-llama/Llama-3.1-8B-Instruct} &
\url{https://ai.meta.com/llama/license} \\
\addlinespace
Qwen1.5-7B-Instruct &
\url{https://huggingface.co/Qwen/Qwen1.5-7B-Chat} &
\url{https://huggingface.co/Qwen/Qwen1.5-7B-Chat/blob/main/LICENSE} \\
\addlinespace
Deepseek-Llama-8B &
\url{https://huggingface.co/deepseek-ai/DeepSeek-R1-Distill-Llama-8B} &
\url{https://huggingface.co/datasets/choosealicense/licenses/blob/main/markdown/mit.md} \\
\addlinespace
Deepseek-Qwen-7B &
\url{https://huggingface.co/deepseek-ai/DeepSeek-R1-Distill-Qwen-7B} &
\url{https://huggingface.co/datasets/choosealicense/licenses/blob/main/markdown/mit.md} \\
\bottomrule
\end{tabularx}
\caption{Models, download sources, and licenses used in our experiments.}
\label{tab:model_details}
\end{table*}

\subsubsection*{Instruction}
\begin{figure}[h!]
    \centering
    \includegraphics[width=1.0\linewidth]{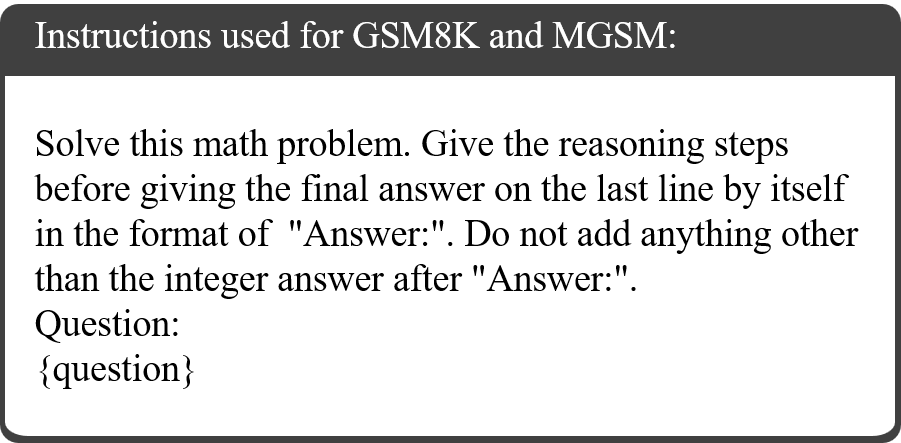} 
    \caption{The instruction used to prompt LLMs for GSM8K and MGSM datasets.}
    \label{fig:gsm8k_mgsm_instruction}
\end{figure}

\begin{figure}[h!]
    \centering
    \includegraphics[width=1.0\linewidth]{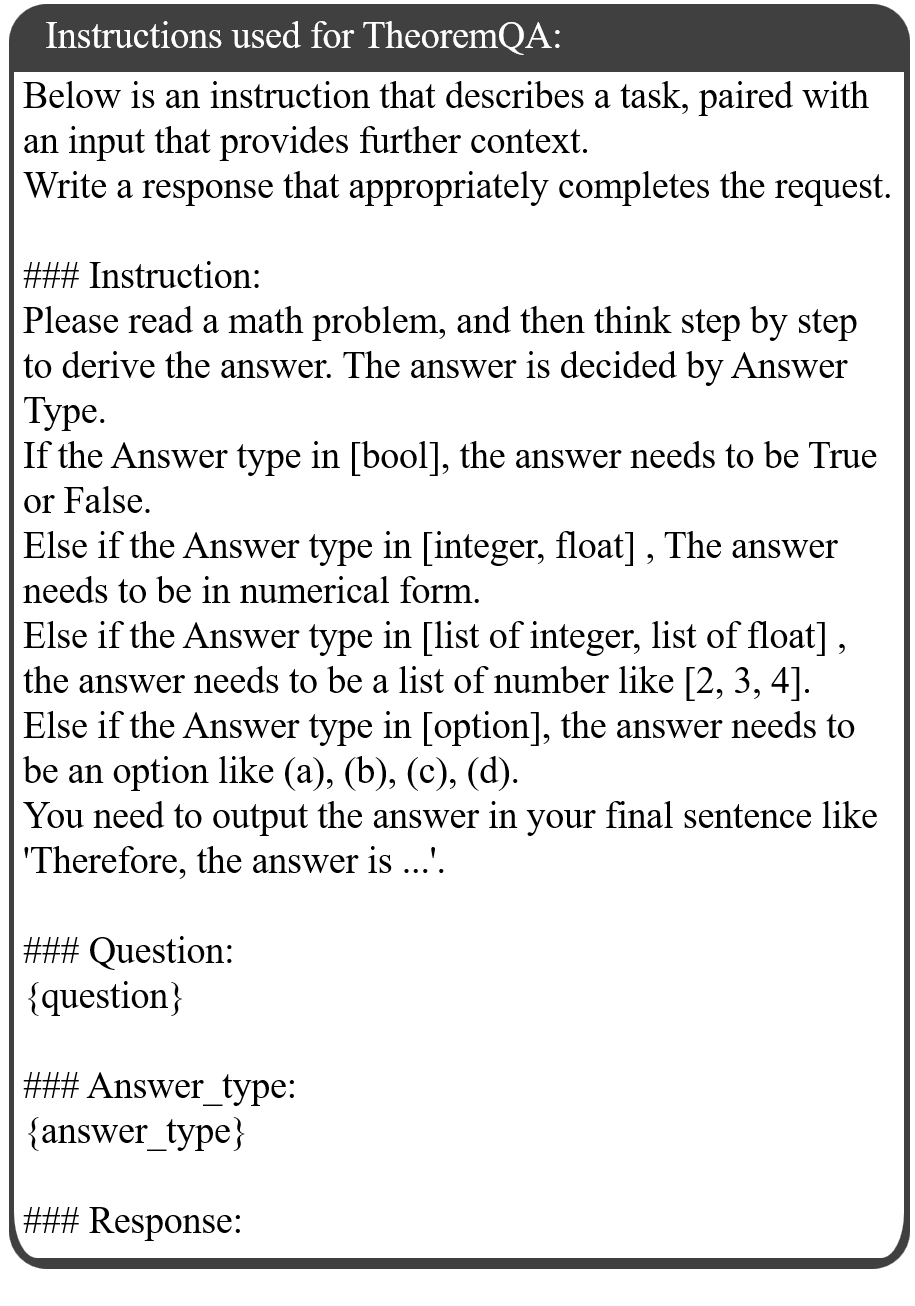} 
    \caption{The instruction used to prompt LLMs for TheoremQA dataset.}
    \label{fig:theoremqa_instruction}
\end{figure}

Following CoE~\cite{wang2025latentspacechainofembeddingenables}, we directly reuse the evaluation instructions provided in their work to ensure consistency and professionalism.Specifically, the instructions used for each dataset are listed in figs  \ref{fig:gsm8k_mgsm_instruction} - fig \ref{fig:theoremqa_instruction}.

\begin{figure}[h!]
    \centering
    \includegraphics[width=1.0\linewidth]{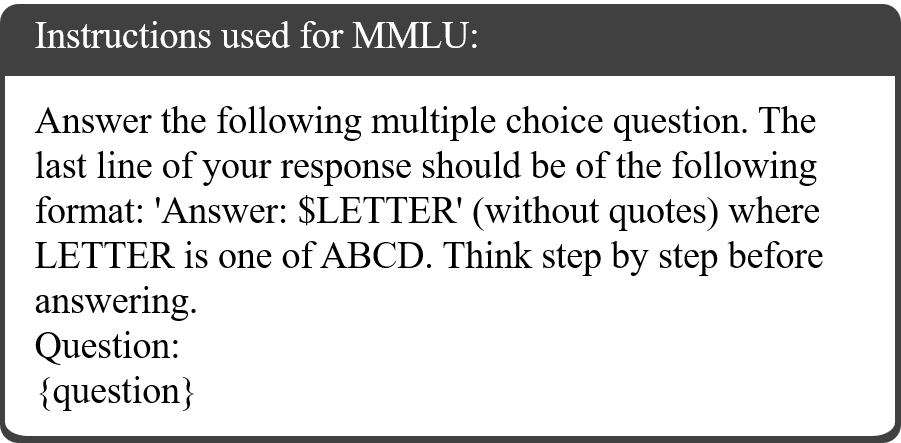} 
    \caption{The instruction used to prompt LLMs for MMLU dataset.}
    \label{fig:mmlu_instruction}
\end{figure}

\begin{figure}[h!]
    \centering
    \includegraphics[width=1.0\linewidth]{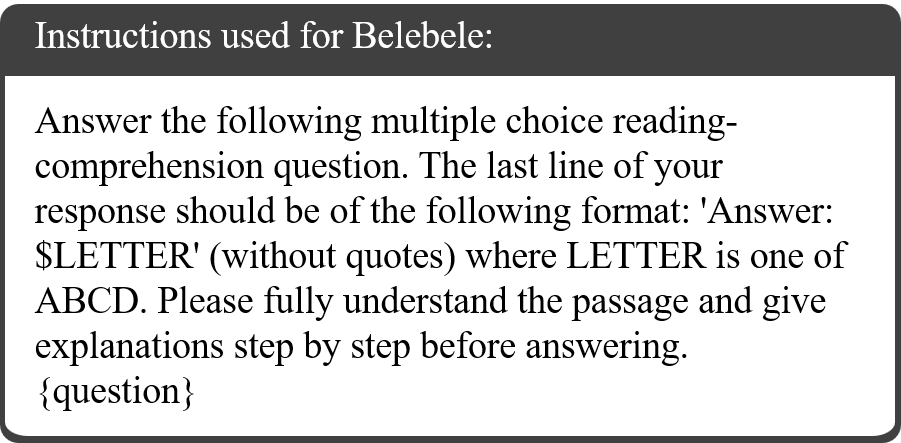} 
    \caption{The instruction used to prompt LLMs for Belebele dataset.}
    \label{fig:belebele_instruction}
\end{figure}

\subsubsection*{Inference}
We set the maximum output length to 1024 tokens and used the \texttt{<eos\_token>} for truncation. The inference process employs greedy decoding without random sampling. All the models we used are 7B/8B, thus in the all experiments we use 2 RTX4090.

\subsubsection*{Baselines}

We present the mathematical definitions of the seven baseline uncertainty metrics used in our evaluation.

\begin{itemize}

  \item \textbf{Maximum Softmax Probability}~\cite{hendrycks2017baseline}   
  
  Reflects the model's confidence by averaging the maximum predicted probability at each decoding step:
  \[
  \text{MaxProb} = \mathbb{E}_{1 \leq t \leq T} [\max \mathbf{p}_t]
  \]

  \item \textbf{Perplexity}~\cite{si2023promptinggpt3reliable}  
  
  Approximates the weighted branching factor of a language by exponentiating the average negative log of the maximum predicted probability:
  \[
  \text{PPL}(x) = \mathbb{E}_{1 \leq t \leq T} [-\log(\max \mathbf{p}_t)]
  \]

  \item \textbf{Entropy}~\cite{Huang_2025}  
  
  Measures the average distributional uncertainty across the sequence:
  \[
  \text{Entropy} = \mathbb{E}_{1 \leq t \leq T} \left[ \mathbb{E} \left[ -\log \mathbf{p}_{t} \right] \right]
  \]

  \item \textbf{Temperature Scaling}~\cite{shih2023long}  
  
  Before applying softmax to logits $\mathbf{z}_t = (z_{t,1}, z_{t,2}, \ldots, z_{t,|V|})$ at each decoding step $t$, 
  we divide them by a temperature parameter $T > 0$ to calibrate the output distribution:
  \[
  \mathbf{p}_t^{(T)} = \text{softmax}\left( \frac{\mathbf{z}_t}{T} \right), \quad \text{i.e.} \quad
  p_{t,i}^{(T)} = \frac{\exp\left(\frac{z_{t,i}}{T}\right)}{\sum_{d=1}^{|V|} \exp\left(\frac{z_{t,d}}{T}\right)}.
  \]
  In our experiments, we set $T=0.7$.  
  Subsequent calculations are consistent with Maximum Softmax Probability:
  \[
  \text{Temperature Scaling} = \mathbb{E}_{1 \leq t \leq T} \left[ \max_i p_{t,i}^{(T)} \right]
  \]

  \item \textbf{Energy}~\cite{liu2020energy}  
  
  Maps logits to an energy value as a substitute for softmax-based confidence. At each step $t$, the energy is computed as:
  \[
  \text{Energy} = \mathbb{E}_{1 \leq t \leq T} \left[ -T \cdot \log \left( \sum_{d=1}^{|V|} \exp\left(\frac{z_{t,d}}{T} \right) \right) \right]
  \]
  We set $T = 0.7$ in our experiments.

  \item \textbf{CoE-R}~\cite{wang2025latentspacechainofembeddingenables}
  
  A confidence score computed by averaging the difference between relative magnitude and relative angle changes across adjacent hidden states.
  \[
  \text{CoE-R}(H) = \frac{1}{L} \cdot \sum_{l=0}^{L-1} \left( \frac{M(h_l, h_{l+1})}{M(h_0, h_L)} - \frac{A(h_l, h_{l+1})}{A(h_0, h_L)} \right)
  \]

  \item \textbf{CoE-C}~\cite{wang2025latentspacechainofembeddingenables}
    
    Let each transition from layer $l$ to $l+1$ be a complex number $C_l$, where its magnitude is $M_{l,l+1}$ and its angle (argument) is $A_{l,l+1}$:
    \[
    C_l = M_{l,l+1} e^{i A_{l,l+1}}
    \]
    The CoE-C score is then the magnitude of the average of all these complex numbers:
    \[
    \text{CoE-C}(H) = \left| \frac{1}{L} \sum_{l=0}^{L-1} C_l \right|
    \]

\end{itemize}

\subsubsection*{Evaluation Metrics}
In this section, we provide the mathematical formulations of the three evaluation metrics used in our study.
\begin{itemize}

    \item \textbf{AUROC (Area Under Receiver Operating Characteristic)}  
    
    Let $\mathcal{S}_\text{pos}$ and $\mathcal{S}_\text{neg}$ be the sets of positive and negative samples, and let $s_i$ denote the confidence score of sample $i$.

    AUROC measures the probability that a randomly chosen positive sample has a higher score than a randomly chosen negative one:
    \begin{equation}
    \text{AUROC} = \frac{1}{|\mathcal{S}_\text{pos}||\mathcal{S}_\text{neg}|} \sum_{i \in \mathcal{S}_\text{pos}} \sum_{j \in \mathcal{S}_\text{neg}} \mathbb{1}(s_i > s_j),
    \end{equation}
    where $\mathbb{1}(\cdot)$ is the indicator function.

    \item \textbf{FPR@95 (False Positive Rate at 95\% Recall)}  
    
    FPR@95 is the false positive rate at the threshold where recall reaches 95\%.

    Let $\tau^\ast$ be the threshold such that:
    \begin{equation}
    \tau^\ast = \arg\min_{\tau} \left| \text{Recall}(\tau) - 0.95 \right|.
    \end{equation}
    Then:
    \begin{equation}
    \text{FPR@95} = \frac{\text{FP}(\tau^\ast)}{|\mathcal{S}_\text{neg}|}.
    \end{equation}

    \item \textbf{AUPR (Area Under Precision-Recall Curve)}  
   
    Let $\tau$ be a threshold. Define:
    \begin{itemize}
        \item $\text{TP}(\tau)$: true positives with $s_i \geq \tau$
        \item $\text{FP}(\tau)$: false positives with $s_i \geq \tau$
        \item $\text{FN}(\tau)$: false negatives with $s_i < \tau$
    \end{itemize}

    Then:
    \begin{align}
    \text{Precision}(\tau) &= \frac{\text{TP}(\tau)}{\text{TP}(\tau) + \text{FP}(\tau)}, \\
    \text{Recall}(\tau) &= \frac{\text{TP}(\tau)}{\text{TP}(\tau) + \text{FN}(\tau)}.
    \end{align}

    AUPR is the area under the precision-recall curve:
    \begin{equation}
    \text{AUPR} = \int_0^1 \text{Precision}(r) \, dr,
    \end{equation}
    where $r = \text{Recall}(\tau)$ varies with the threshold. This is computed via numerical integration.

\end{itemize}


\section*{D. Additional Figures}
\begin{figure}[h!]
    \centering
    \includegraphics[width=1.0\linewidth]{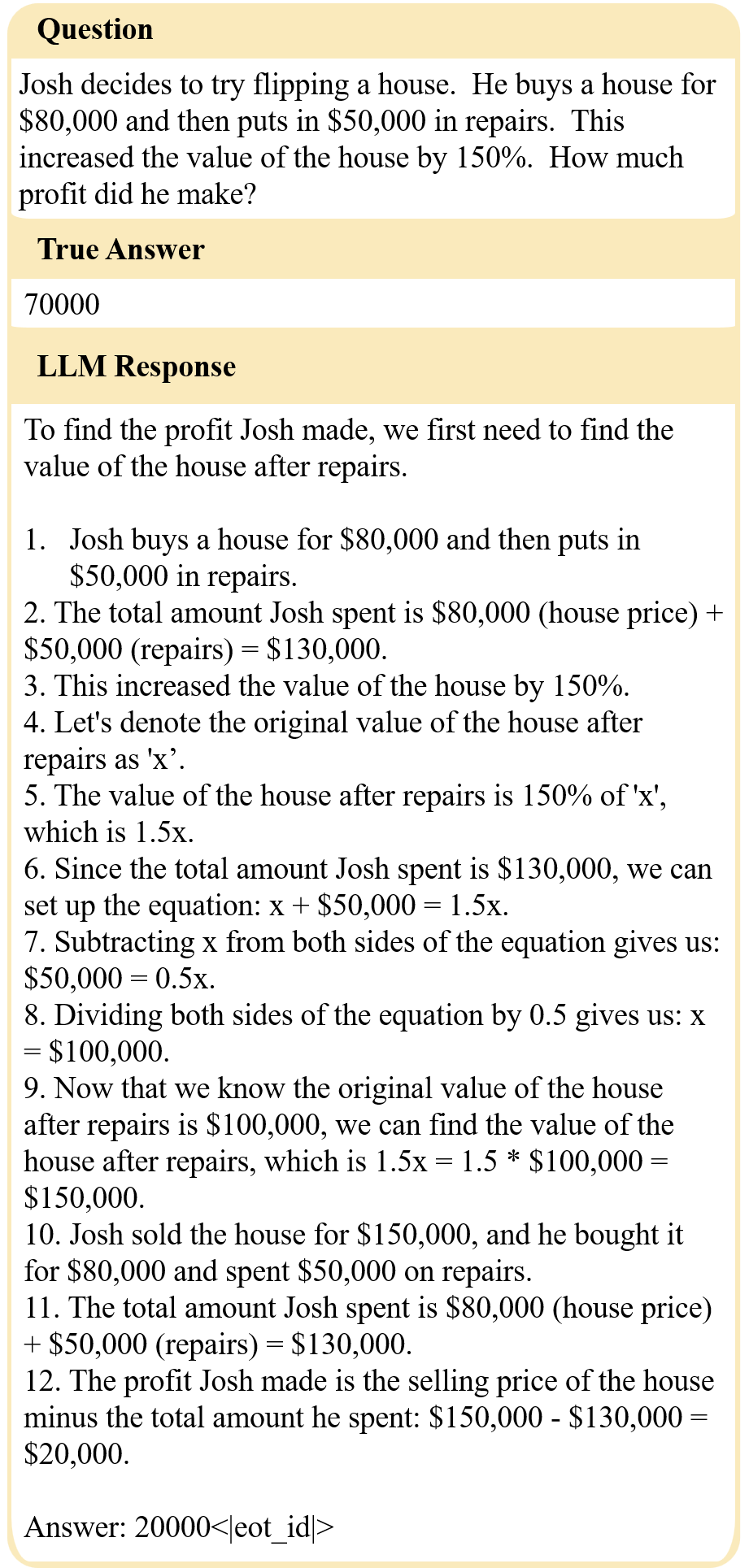} 
    \caption{The hallucinated sample 1 for token representation visualization.}
    \label{fig:hallucinated_example}
\end{figure}

\begin{figure}[h!]
    \centering
    \includegraphics[width=1.0\linewidth]{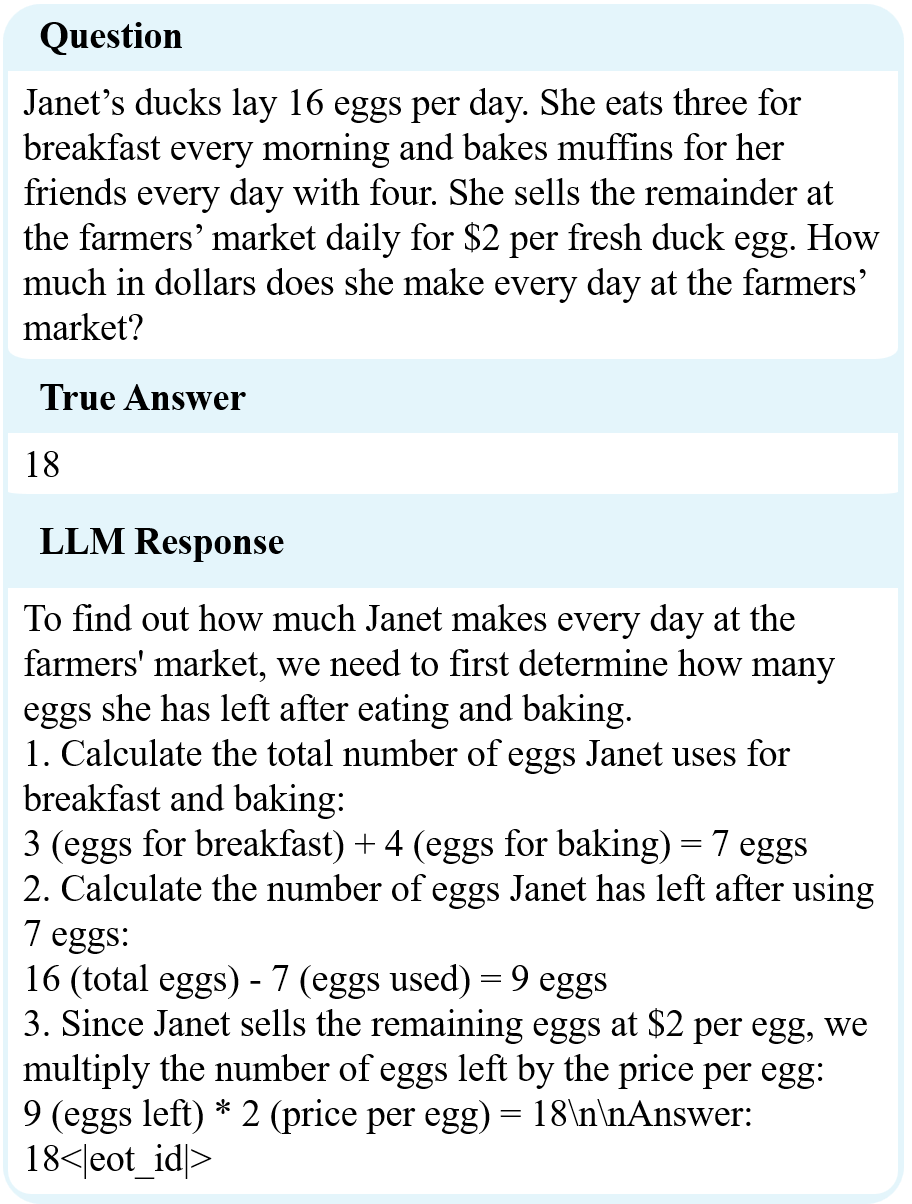} 
    \caption{The faithful sample 1 for token representation visualization.}
    \label{fig:faithful_example}
\end{figure}

\begin{figure}[h!]
    \centering
    \includegraphics[width=1.0\linewidth]{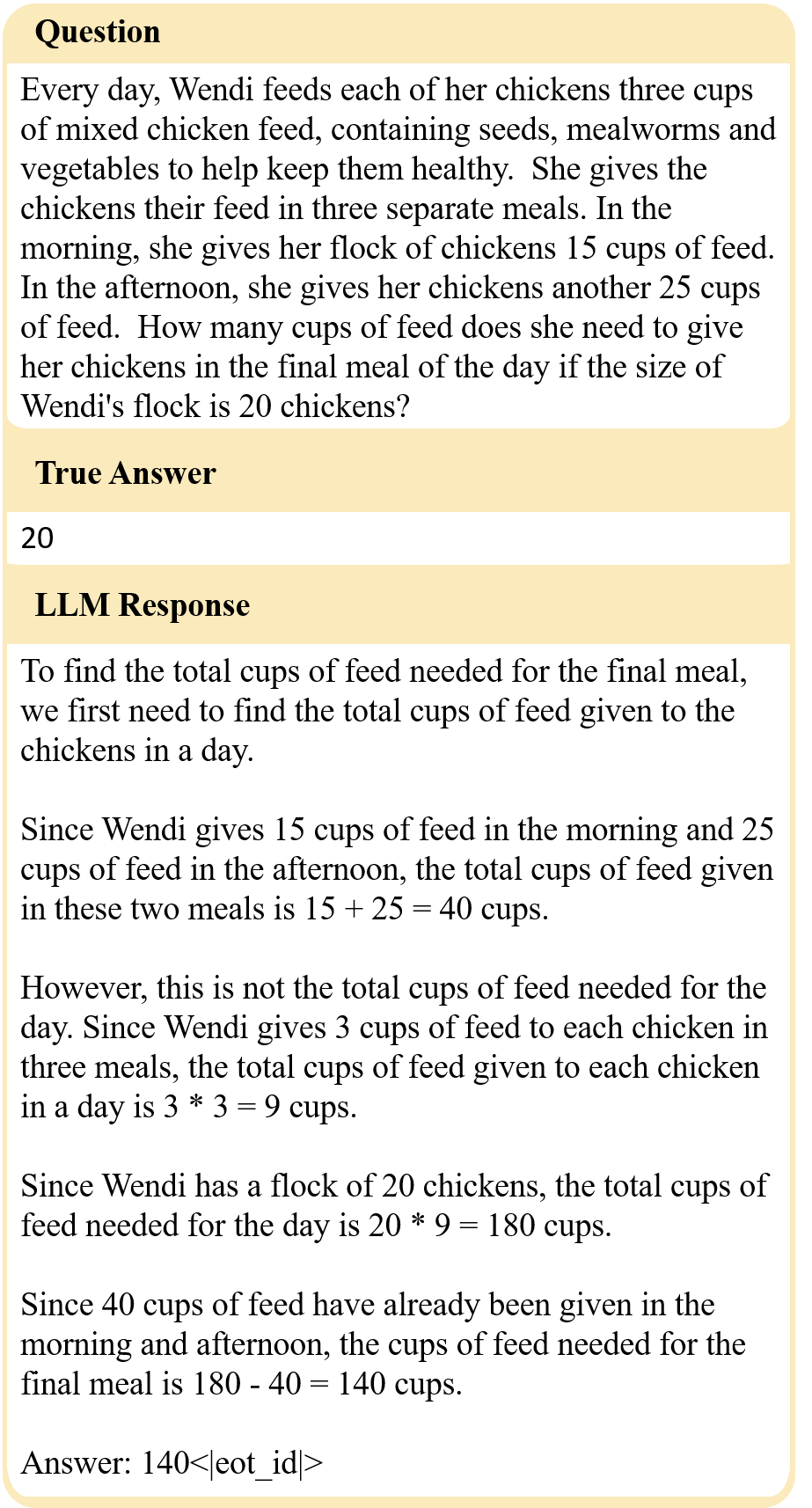} 
    \caption{The hallucinated sample 2 for token representation visualization.}
    \label{fig:hallucinated_example_1}
\end{figure}

\begin{figure}[h!]
    \centering
    \includegraphics[width=1.0\linewidth]{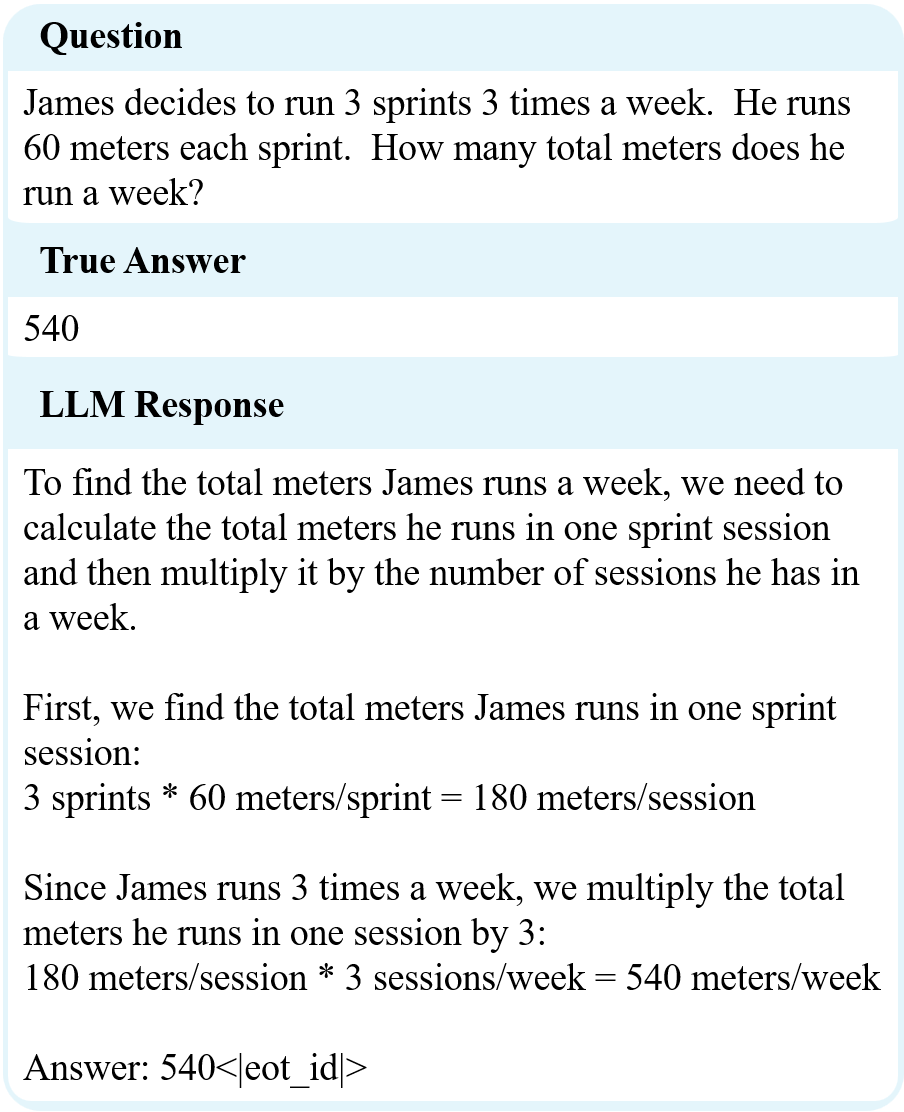} 
    \caption{The faithful sample 2 for token representation visualization.}
    \label{fig:faithful_example_1}
\end{figure}

\begin{figure}[h!]
    \centering
    \includegraphics[width=1.0\linewidth]{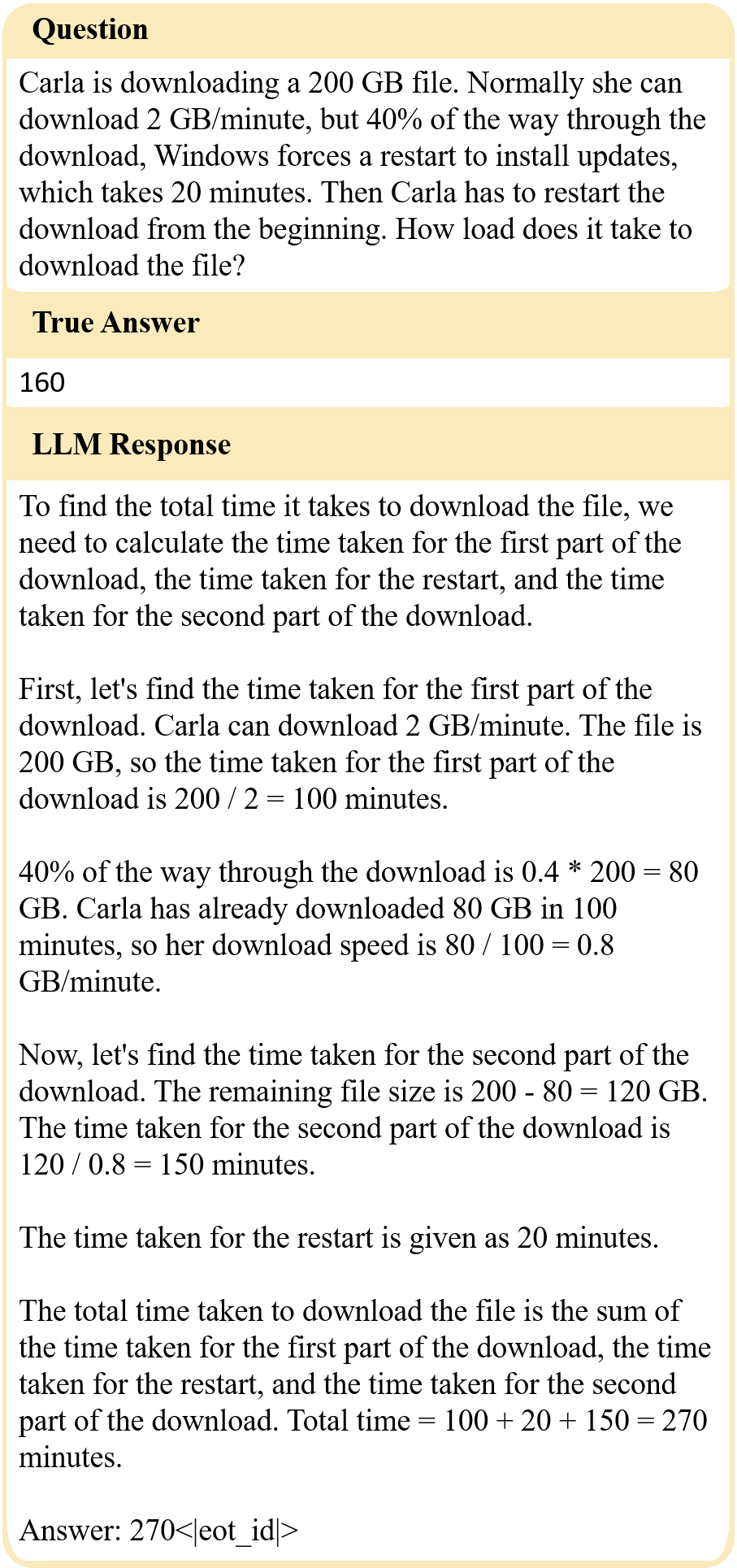} 
    \caption{The hallucinated sample 3 for token representation visualization.}
    \label{fig:hallucinated_example_2}
\end{figure}

\begin{figure}[h!]
    \centering
    \includegraphics[width=1.0\linewidth]{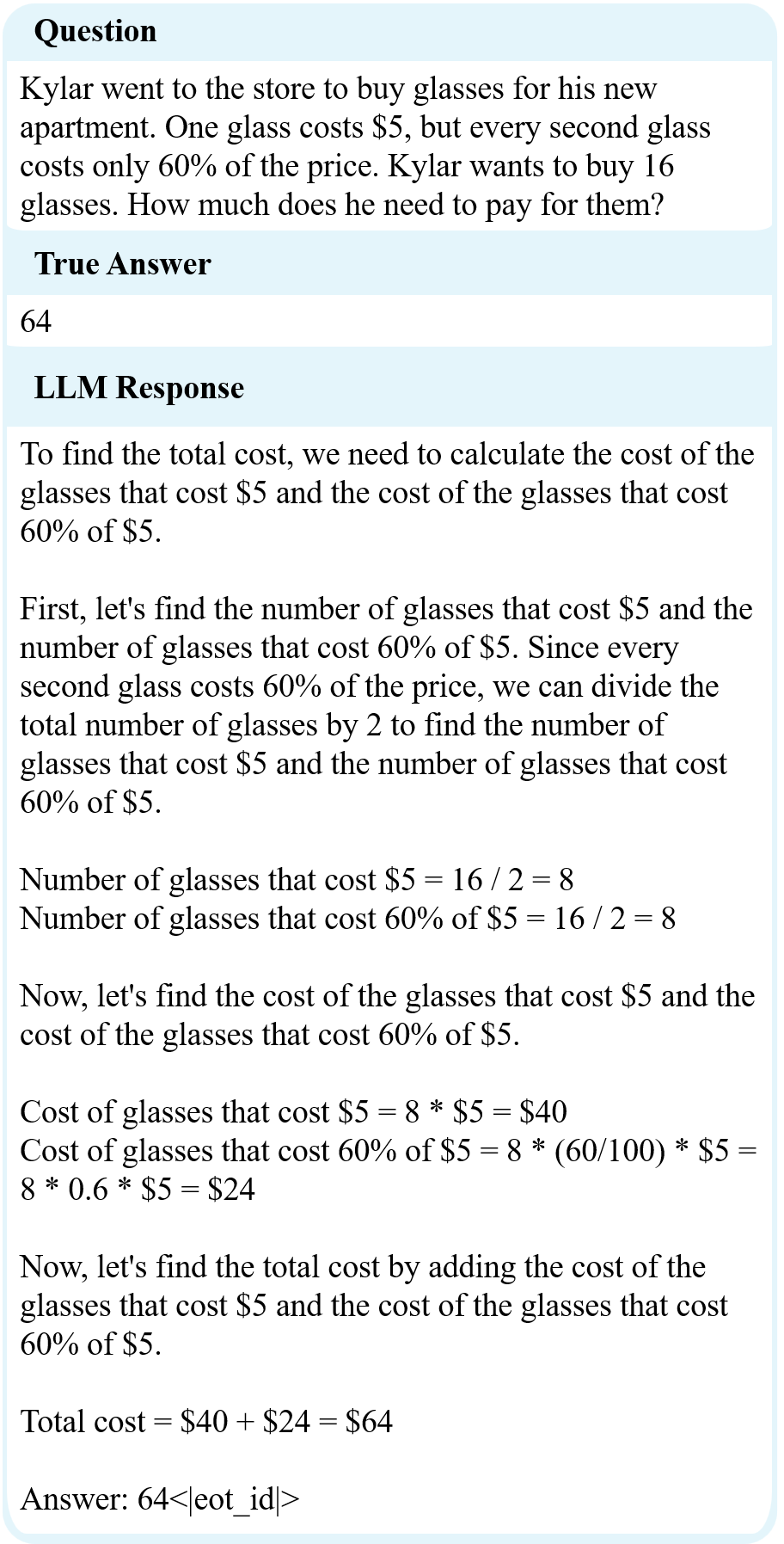} 
    \caption{The faithful sample 3 for token representation visualization.}
    \label{fig:faithful_example_2}
\end{figure}

\begin{figure}[h!]
    \centering
    \includegraphics[width=1.0\linewidth]{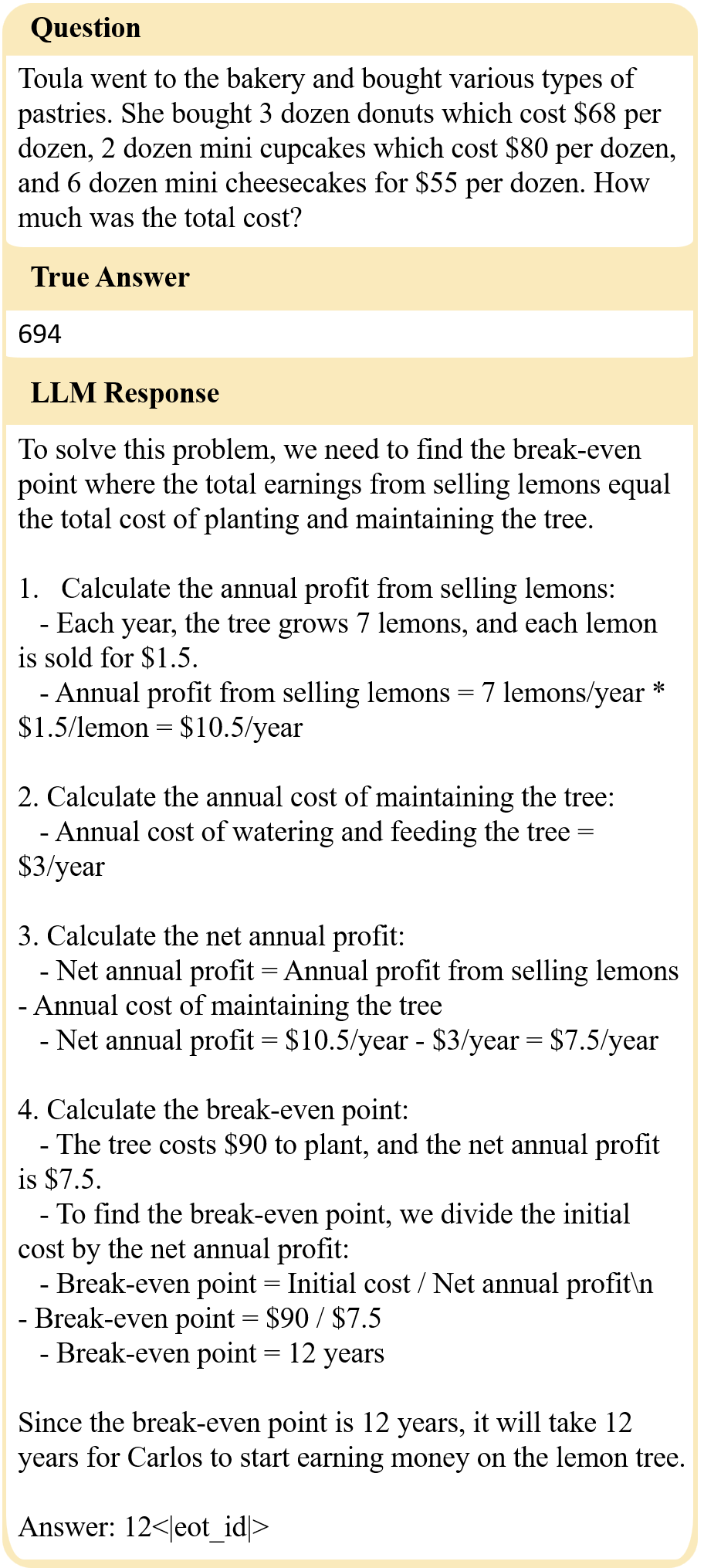} 
    \caption{The hallucinated sample 4 for token representation visualization.}
    \label{fig:hallucinated_example_3}
\end{figure}

\begin{figure}[h!]
    \centering
    \includegraphics[width=1.0\linewidth]{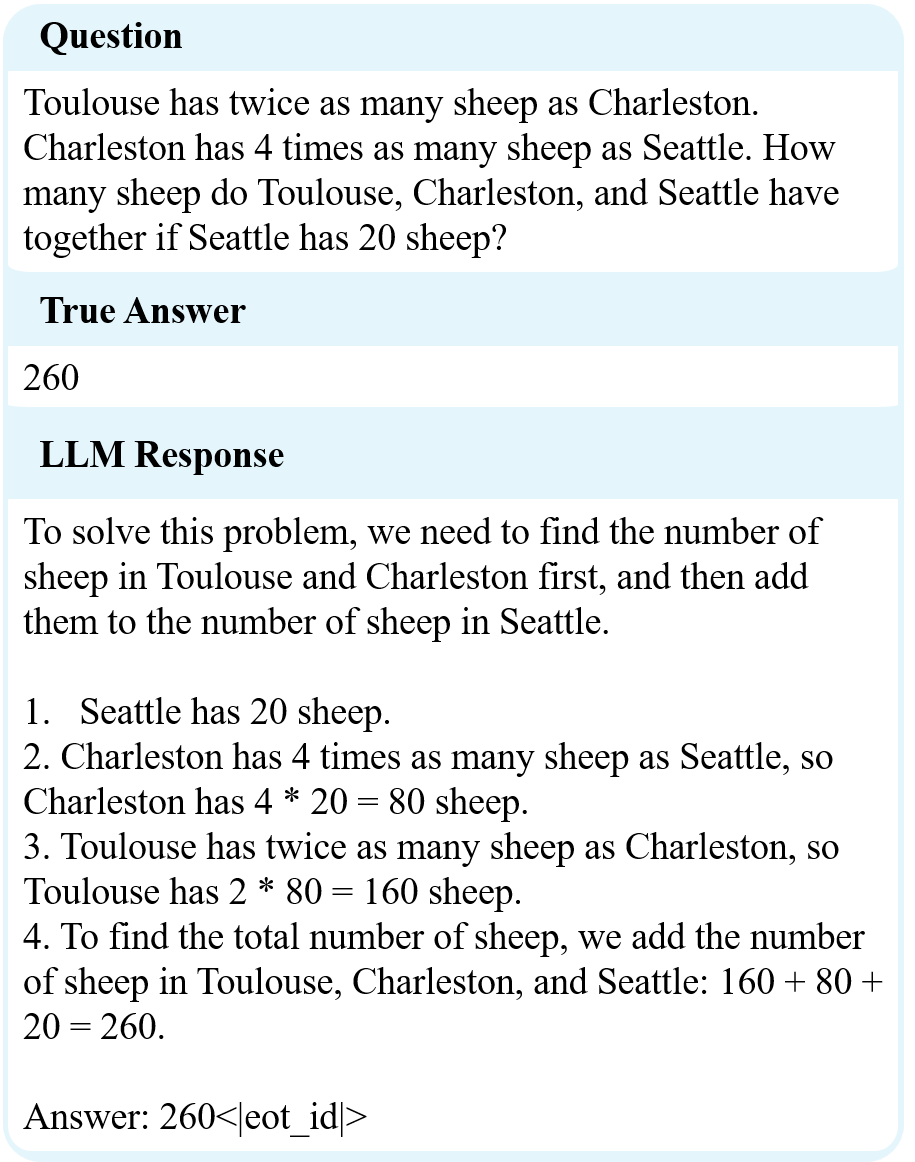} 
    \caption{The faithful sample 4 for token representation visualization.}
    \label{fig:faithful_example_3}
\end{figure}

To support the visualization of token dispersion patterns in the Method Section, we present some representative examples used in our dispersion's visualization analysis: as shown in fig \ref{fig:hallucinated_example} - \ref{fig:faithful_example_3}.

\section*{E. Limitation and Future Work}
A primary limitation of our approach is its nature as a "white-box" method, which requires access to internal hidden states and thus precludes its use on closed-source models like GPT-4. However, we argue this is a necessary trade-off for achieving deeper analytical insight and greater efficiency.

Black-box methods, while universally applicable, often come with significant costs. As previously discussed, they typically rely on computationally expensive techniques such as generating multiple samples or making calls to external tools, which can be both slow and resource-intensive. Furthermore, their analysis remains indirect, inferring model behavior from output patterns alone without revealing the underlying causes.

In contrast, white-box approaches like ours offer a more direct and efficient path. By analyzing internal representations, our method provides a higher degree of interpretability, revealing the mechanistic underpinnings of the generation process. This makes the white-box paradigm fundamentally more aligned with the scientific goal of not just detecting, but truly understanding and improving large language models. Therefore, while black-box techniques provide a practical solution for proprietary APIs, we contend that white-box methods offer a more principled and insightful avenue for advancing the field.

\end{document}